\documentclass[pdflatex,sn-mathphys]{sn-jnl}
\theoremstyle{thmstyleone}%
\usepackage{pdflscape}
\usepackage{array}
\usepackage{caption}
\usepackage{dirtytalk}
\usepackage{parskip}
\usepackage{blindtext}
\usepackage{amsmath}
\usepackage{natbib}
\usepackage{booktabs} 
\usepackage{float}    

\usepackage{graphicx} 
\usepackage{float}    
\usepackage{subcaption} 
\usepackage{booktabs} 
\usepackage{adjustbox} 
\usepackage{geometry} 

\usepackage{amsmath}    
\usepackage{amssymb}    
\usepackage{mathpazo}   

\theoremstyle{thmstyletwo}%
\theoremstyle{thmstylethree}%

\usepackage{booktabs} 
\usepackage{soul}
\usepackage{float}
\usepackage{amsmath,lipsum}
\usepackage[english]{babel}
\usepackage{csquotes}
\usepackage{appendix}

\begin{document}
\title[Machine Learning and Deep Learning for Anomaly Detection]{Unsupervised  anomaly detection in large-scale estuarine acoustic telemetry data}
\author[1,2]{Siphendulwe Zaza}
\author*[1,2]{Marcellin Atemkeng}
\author[2,3]{Taryn S. Murray}
\author[2]{John David Filmalter}
\author[2]{Paul D. Cowley}
\affil[1]{\small Department of Mathematics, Rhodes University, PO Box 94, Makhanda, 6140, South Africa}
\affil[2]{\small South African Institute for Aquatic Biodiversity, {\textcolor{black}{Private Bag 1015}}, Makhanda, 6140, South Africa}
\affil[3]{\small Department of Ichthyology and Fisheries Science, Rhodes University, PO Box 94, Makhanda, 6140, South Africa}
\affil[*]{\small Correspondence: m.atemkeng@ru.ac.za}

\abstract{
{\textcolor{black}{
Acoustic telemetry data plays a vital role in understanding the behaviour and movement of aquatic animals. However, these datasets, which often consist of millions of individual data points, frequently contain anomalous movements that pose significant challenges. Traditionally, anomalous movements are identified either manually or through basic statistical methods, approaches that are time-consuming and prone to high rates of unidentified anomalies in large datasets.
This study focuses on the development of automated classifiers for a large telemetry dataset comprising detections from fifty acoustically tagged dusky kob (“\textit{Argyrosomus japonicus}”) monitored in the Breede Estuary, South Africa. Using an array of 16 acoustic receivers deployed throughout the estuary between 2016 and 2021, we collected over three million individual data points.
We present detailed guidelines for data pre-processing, resampling strategies, labelling process, feature engineering, data splitting methodologies, and the selection and interpretation of machine learning and deep learning models for anomaly detection. Among the evaluated models, neural networks autoencoder (NN-AE) demonstrated superior performance, aided by our proposed threshold-finding algorithm. NN-AE achieved a high recall with no false normal (i.e., no misclassifications of anomalous movements as normal patterns), a critical factor in ensuring that no true anomalies are overlooked. In contrast, other models exhibited false normal fractions exceeding 0.9, indicating they failed to detect the majority of true anomalies—a significant limitation for telemetry studies where undetected anomalies can distort interpretations of movement patterns.
While the NN-AE’s performance highlights its reliability and robustness in detecting anomalies, it faced challenges in accurately learning normal movement patterns when these patterns gradually deviated from anomalous ones. To the best of our knowledge, this study represents the first effort to develop automated methods leveraging machine learning and deep learning to address anomalous detections in acoustic telemetry data.}}}

\keywords{{\textcolor{black}{Acoustic telemetry, \textit{Argyrosomus japonicus}, machine learning, deep learning, Breede Estuary, dusky kob}}}



\maketitle
\section{Introduction}
{\textcolor{black}{Studying fish movement and behaviour is crucial for understanding ecological dynamics and implementing effective conservation strategies \cite{crossin2017acoustic, lowerre2019ocean}. Movements can be studied using various methods, including conventional dart tags providing low-resolution, coarse-scale data \cite{hughes2022movement}, and acoustic telemetry which provides high-resolution, fine-scale data \cite{thorstad2013electronic, hussey2015aquatic}. Acoustic telemetry, which involves the recording of an acoustic signal released from an acoustic transmitter on a deployed acoustic receiver, is currently the most popular method for studying the movements of aquatic animals, providing valuable insights into their movements, habitat use, and survival \cite{hussey2015aquatic, matley2022global}.}

One of the biggest strengths of acoustic telemetry is the amount of data collected, which in some instances can reach millions of individual data points of numerous individuals from various species \cite{dhellemmes2023atfiltr}. However, this is also one of its biggest challenges. Large datasets have their own difficulties when it comes to analyses \cite{simpfendorfer2015ghosts}. Associated with (but not limited to) large datasets are anomalous or false detections \cite{simpfendorfer2015ghosts, dhellemmes2023atfiltr}. These anomalies can arise for various reasons, and occur as a result of a collision of acoustic transmissions which creates a detection of a different transmitter ID by the acoustic receiver \cite{simpfendorfer2015ghosts}. These anomalies can significantly skew the interpretation of fish movements and behaviour patterns \cite{chambert2015modeling, simpfendorfer2015ghosts}, making it essential to develop a strong anomaly detection system to ensure data integrity. In smaller telemetry datasets, anomalies can often be identified visually after plotting the data. However, this method is far too time-consuming for larger datasets comprising millions of detections.

Several statistical packages have been developed that filter, process and analyse passive acoustic telemetry data in the statistical environment R \cite{rcoreteam2022language}, including \textit{actel} \cite{flavio2021actel}, \textit{rsp} \cite{niella2020refined}, \textit{telemetR} \cite{spaulding2024filter} and \textit{ATfiltR} \cite{dhellemmes2023atfiltr}. While each has its strengths, another option to process large quantities of detection data is through machine learning. Machine learning (ML) and deep learning (DL) models are increasingly being incorporated into biological studies. For example, automating the detection of Hainan gibbon \textit{Nomascus hainanus} calls using deep neural networks, and monitoring population numbers of wildebeest and zebra using a U-Net-based deep learning model integrated with a post-processing clustering model \cite{Wu2023deep}. In the aquatic environment, ML and DL models have been applied to underwater video analysis to detect atypical fish behaviour \cite{wang2020anomalous}, and while their potential in acoustic telemetry remains relatively underexplored, these models are slowly being incorporated into telemetry studies. For example, the migration fate of Atlantic salmon \textit{Salmo salar} smolts was classified using unsupervised k-means cluster analyses and supervised random forest models \cite{notte2022application}, predatory fishes have been identified through the use of predation tags and associated movements using random forest algorithms \cite{klinard2020application},  relative habitat selection by multiple shark species was predicted using random forest models \cite{griffin2021novel}, and habitat suitability of two estuarine species was modeled using deep feed-forward Artificial Neural Networks \cite{guénard2020deep}. 

Acoustic telemetry data, in essence, collects presence/absence data of tagged individuals. These data can be reduced to a time series, listing the dates and times tagged individuals (which have been tagged with acoustic transmitters that have unique ID codes) and the location at which the detections occurred. Traditional ML methods, such as Density-Based Spatial Clustering of Applications with Noise (DBSCAN), Local Outlier Factor (LOF), and Isolation Forest (IF), have been widely used to detect anomalies within time series data \cite{schindler2023towards}. The effectiveness of DL approaches such as Neural Network Autoencoder (NN-AE) in detecting anomalies has also been shown \cite{chen2021anomaly}. However, a major limitation of traditional methods is their inability to capture complex temporal dependencies in time series data, which can lead to poor performance in certain scenarios \cite{benova2024comprehensive}. In the context of big data, where traditional statistical methods may fall short, ML and DL classifiers, which can significantly improve the accuracy of false detections, offer a robust alternative for ensuring data integrity. This paper addresses the gap by detailing the training of ML and DL classifiers for the accurate identification of false detections of an important estuary-dependent species, dusky kob \textit{Argyrosomus japonicus}, within a permanently open estuary along South Africa's southern coast. This paper aims to provide practitioners with a comprehensive understanding of the training process, the decision-making involved in model selection, and practical guidelines for these decisions.

\section{Materials and methods}\label{Materials and methods}
\subsection{Study site and species}
Fish were tagged in the large permanently open Breede Estuary situated in the Western Cape Province, South Africa (Figure \ref{Study_Site }). The estuary is 52 km in length and is a vital ecological and hydrological system that flows into the Indian Ocean at the coastal town of Witsand \cite{ziko2023acoustic}. The Breede Estuary features diverse habitats like mudflats, channels, and intertidal zones, supporting a wide range of aquatic species, including several fish species of recreational and small-scale/subsistence importance \cite{childs2015habitat}, such as spotted grunter \textit{Pomadasys commersonnii}, leervis \textit{Lichia amia} and dusky kob \cite{cowley2006space}. 

\begin{figure}
    \centering
    \includegraphics[width=1.\textwidth]{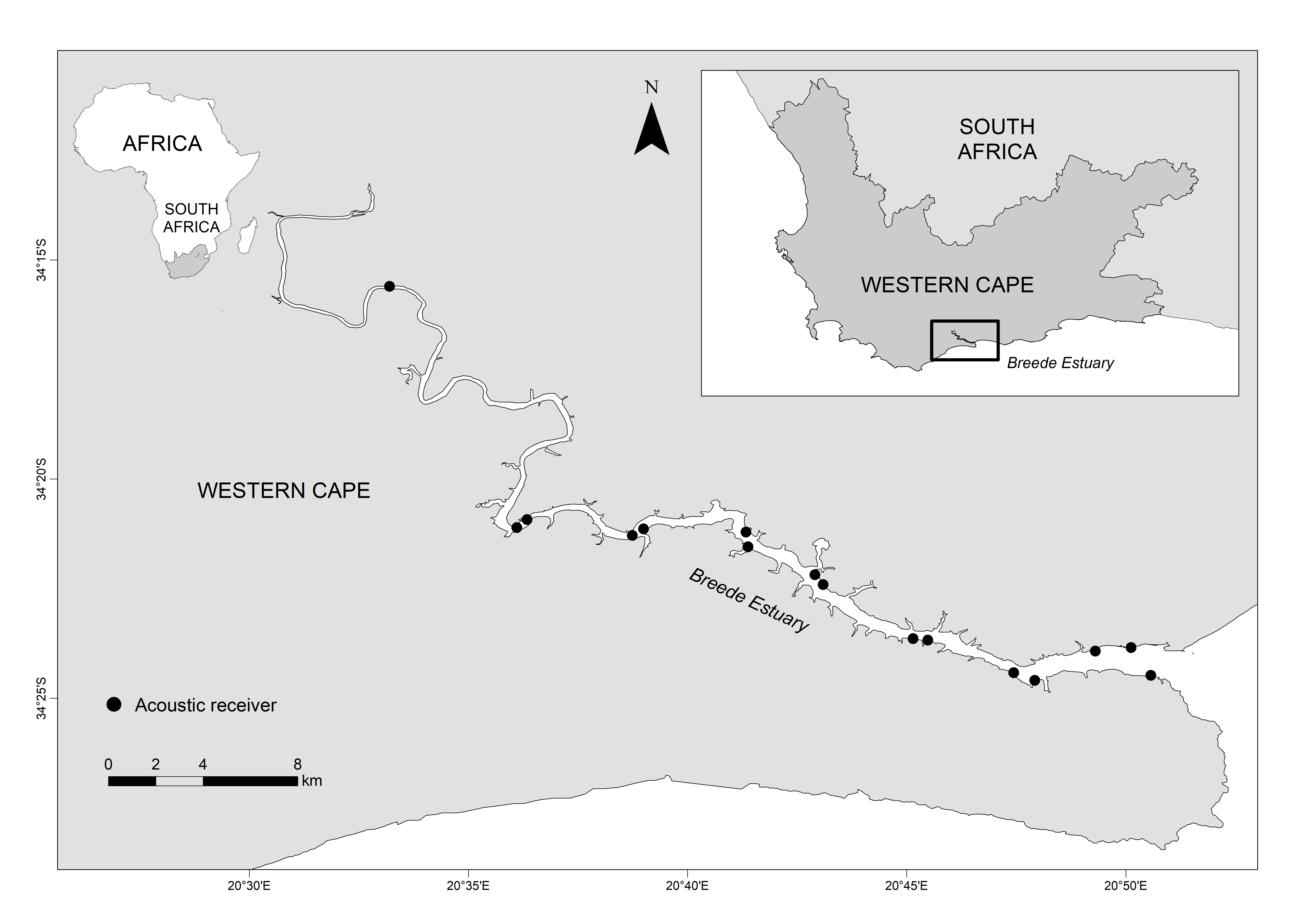}
    \caption{{Map of the Breede Estuary showing the locations of acoustic receivers (black pins) deployed to monitor dusky kob movements in the estuary between 2016 and 2021.}}
    \label{Study_Site }
\end{figure}

The dusky kob is an important coastal fishery species with a South African distribution along the southeast coast from Cape Point to southern Mozambique (more abundant from Cape Agulhas to northern KwaZulu-Natal) \cite{griffiths1995contribution}. Its life history is well known, with juveniles being dependent on estuaries as nursery habitats \cite{griffiths1996life, termorhuizen1996influence}, and even large adults remaining in estuaries for extended periods of time (up to seven or more years; ATAP, unpublished data). Their movements in South Africa have been well studied using acoustic telemetry \cite{cowley2008estuarine, naesje2012riding, childs2013estuary, childs2015habitat}. Juveniles are highly resident to estuaries \cite{cowley2008estuarine, childs2015habitat}, and as adults (ATAP, unpublished data), making extensive use of these systems while in them \cite{naesje2012riding}. Given the long-term residency to estuaries, and their wide-scale use of most estuaries in which they occur (i.e. lower to upper reaches), acoustic telemetry data of dusky kob collected in estuaries form the ideal time series dataset for ML and DL model training. 

\subsection{Tagging and data collection}
Between 2016 and 2021, 50 dusky kob, ranging in length from 660 to 1720 mm total length (TL) (average ± SD: 1196 ± 279 mm TL) were caught and tagged with individually coded acoustic transmitters (V16 transmitters, 69 kHz, Innovasea, Halifax, Canada) (Table 1). The tagging process involved catching fish using conventional rod and line techniques from a small boat or the bank of the estuary. Once caught, the fish was carefully pulled onto the boat, and placed into a flexible PVC sling, filled with water, and prepared for surgery. The head was covered with a cloth to reduce stress, and the gills remained submerged throughout the surgical procedure. The surgical procedure involved making a small incision (approximately 1.5 to 2 cm) in the fish's abdomen between the pelvic and anal fin, inserting the transmitter into the coelomic cavity, and then stitching the incision with two or three sutures. Once sutured, an antibacterial wound powder was placed over the incision site which congealed upon wetting. The surgical procedure followed that of \cite{cowley2008estuarine}. Ethical approval was obtained from the South African Institute for Aquatic Biodiversity's Animal Ethics Committee (approval no. $2013\_06$), and research permits were obtained from the Department of Forestry, Fisheries and the Environment (permit no. RES2017/26, RES2019/16, RES2020/45).

The movements of the tagged dusky kob were monitored in the Breede Estuary between 2016 and 2021 using an array of 16 passive acoustic receivers (model VR2W, 69 kHz, Innovasea, Halifax, Canada) deployed throughout the estuary (Figure \ref{Study_Site }). Receivers were downloaded at least once a year to secure data using Innovasea’s VUE software.

Overall, all 50 dusky kob were detected at least once, together accumulating $3013930$ detections (Table  \ref{tab:my_label1}). The average number of detections recorded per individual was 60278 ($\pm$ 63276), ranging from 23 to 259676 detections, and individuals were detected for an average of 184 ($\pm$ 217) days throughout the monitoring period, ranging from 2 to 1061 days (Table \ref{tab:my_label1}). The average number of detections recorded per day ranged from 11 to 854 average $\pm$ SD: 336 $\pm$ 128). 
\begin{table}
    \centering
    \captionsetup{list=no} 
    \caption{Summary of tagging and acoustic monitoring details for dusky kob (\emph{Argyrosomus japonicus}) in the Breede Estuary (2016-2021), showing total length, tagging date, detection count, station, and monitoring duration for each individual.}
    \begin{tabular}{|p{2.3cm}|p{1.5cm}|p{1.5cm}|p{1.5cm}|p{1.5cm}|p{1.5cm}|}
    \hline
        FishID & Total length (mm TL) &Tag date&No. of detections & No. of unique stations visited & No. of days detected \\ \hline \hline 
        A69-9001-23701 & 1280 & 2015-11-01 & 71148 &15 & 194 \\ \hline
        A69-9001-23702 & 990 & 2016-01-03 & 23 &1 & 2 \\ \hline
        A69-9001-23668 & 1450 & 2016-10-05 & 42408 &15 & 80 \\ \hline
        A69-9001-23669 & 1600 & 2016-10-15 & 67564 &16 & 164 \\ \hline
        A69-9001-23670 & 1280 & 2016-10-16 & 88973 &15 & 258 \\ \hline
        A69-9001-23671 & 1380 & 2016-10-16 & 76810 & 16 & 226 \\ \hline
        A69-9001-23672 & 1400 & 2016-10-17 & 8927 &14 & 25 \\ \hline
        A69-9001-23728 & 1210 & 2016-10-17 & 73156 & 16 & 250 \\ \hline
        A69-9001-23727 & 1210 & 2016-10-18 & 12865 & 15 & 53 \\ \hline
        A69-9001-23726 & 1460 & 2016-10-18 & 27984 & 15 & 76 \\ \hline
        A69-9001-23737 & 1510 & 2016-10-18 & 84419 & 15 & 249 \\ \hline
        A69-9001-23738 & 1400 & 2016-10-18 & 7905 &14 & 24 \\ \hline
        A69-9001-23739 & 1390 & 2016-10-18 & 57754 & 16 & 188 \\ \hline
        A69-9001-23747 & 1460 & 2016-10-18 & 86081 & 16 & 208 \\ \hline
        A69-9001-23745 & 1180 & 2016-10-19 & 33178 &16 & 134 \\ \hline
        A69-9001-23746 & 1380 & 2016-10-19 & 25535 & 15 & 60 \\ \hline
        A69-9001-23748a & 1140 & 2016-10-19 & 39614 &15 & 123 \\ \hline
        A69-9001-23749 & 1290 & 2016-10-19 & 37648 & 15 & 100 \\ \hline
        A69-9001-23750 & 1239 & 2016-10-19 & 112064 &16 & 276 \\ \hline
        A69-9001-23751 & 1270 & 2016-10-19 & 57073 & 15 & 137 \\ \hline
        A69-9001-23754a & 1180 & 2016-10-20 & 14804 &15 & 63 \\ \hline
        A69-9001-23754b & 705 & 2018-08-10 & 422 & 3 & 23 \\ \hline
        A69-9001-23675b & 750 & 2018-08-10 & 149205 &15 & 542 \\ \hline
        A69-9001-23766b & 1140 & 2016-10-20 & 55565 &16 & 150 \\ \hline
        A69-9001-23753 & 1340 & 2016-10-20 & 29121 &16 & 100 \\ \hline
        A69-9001-23697b & 1230 & 2016-10-20 & 50402 &15 & 122 \\ \hline
    \end{tabular}
    \label{tab:my_label1}
\end{table}
\addtocounter{table}{-1}
\begin{table}
\renewcommand\thetable{\ref{tab:my_label1}}
    \centering
    \captionsetup{list=no} 
    \caption{Cont.}
    \begin{tabular}{|p{2.3cm}|p{1.5cm}|p{1.5cm}|p{1.5cm}|p{1.5cm}|p{1.5cm}|}
    \hline
        FishID &Total length (mm TL)&Tag date &No. of detections & No. of unique stations visited & No. of days detected \\ \hline \hline
        A69-9001-23698b & 1210 & 2016-10-29 & 41306 &15 & 112 \\ \hline
        A69-9001-23682b & 1140 & 2016-10-29 & 15749 &15 & 45 \\ \hline
        A69-9001-50373b & 1290 & 2016-10-29 & 10480 &15 & 54 \\ \hline
        A69-9001-23774 & 1490 & 2016-10-29 & 13504 & 12 &28 \\ \hline
        A69-9001-23741b & 1310 & 2016-10-29 & 80670 & 14 & 212 \\ \hline
        A69-9001-23721 & 1230 & 2016-11-02 & 56559 &16 & 193 \\ \hline
        A69-9001-23722 & 1190 & 2016-11-02 & 259676 & 1 & 1061 \\ \hline
        A69-9001-23723a & 1340 & 2016-11-02 & 9837 &15 & 34 \\ \hline
        A69-9001-23777 & 1190 & 2016-11-02 & 15067 & 13 & 54 \\ \hline
        A69-9001-23778 & 1190 & 2016-11-02 & 29839 &16 & 126 \\ \hline
        A69-9001-23779 & 1160 & 2016-11-02 & 4461 &11 & 18 \\ \hline
        A69-9001-23775a & 1240 & 2016-11-02 & 57840 &16 & 216 \\ \hline
        A69-9001-23724 & 1460 & 2016-11-11 & 2567 &11 & 10 \\ \hline
        A69-9001-23732 & 1720 & 2017-04-12 & 127797 & 16 & 247 \\ \hline
        A69-9001-23769 & 1650 & 2017-05-05 & 4271 & 3 & 5 \\ \hline
        A69-9001-23730 & 1220 & 2017-02-24 & 8094 &15 & 31 \\ \hline
        A69-9001-23858a & 695 & 2017-08-25 & 239874 & 16 & 461 \\ \hline
        A69-9001-23859a & 660 & 2017-08-25 & 17679 &15 & 63 \\ \hline
        A69-9001-23860a & 665 & 2017-08-26 & 214246 & 15 & 561 \\ \hline
        A69-9001-23861a & 660 & 2017-08-30 & 44924 &10 & 135 \\ \hline
        A69-9001-23862 & 680 & 2017-09-19 & 225462 &16 & 953 \\ \hline
        A69-9001-23714a & 700 & 2017-11-28 & 16364 &9 & 37 \\ \hline
        A69-9001-23684b & 695 & 2018-03-29 & 96051 &15 & 484 \\ \hline
        A69-9001-23733 & 665 & 2018-09-20 & 110965 &14 & 251 \\ \hline
    \end{tabular}
    \label{tab:my_label2}
\end{table}

\subsection{Data pre-processing}
Prior to training ML and DL classifiers, several pre-processing steps were necessary, including cleaning the data to remove noise, normalising the data to ensure consistency, dealing with missing samples, and splitting the dataset into training, validation, and test sets for training and evaluation \cite{marciuc2021data}.

\subsection{Engineered features}
Engineered features are a fundamental concept in ML and DL. They involve creating new features that are specifically informative for the task at hand, with the aim of improving the model's performance \cite{zheng2018feature}. In this study, new features were introduced to capture details not directly present in the data. These included \say{duration at the same station},  \say{number of detections}, \say{number of days detected}, \say{number of unique stations}, and \say{consecutive missing stations} (Table \ref{Features_description3}).  The feature \say{duration at the same station} was of crucial importance as it identified cases where a fish was recorded at only one station during the entire study, possibly indicating a malfunction of the tag or a mortality. The \say{number of detections} and \say{number of days recorded} features provided information about the fish's activity level and habitat use. The feature \say{number of unique stations} provided information about the spatial extent of the fish movements, and \say{consecutive missing stations} indicated possible gaps in the movements. Additionally, the feature \say{distance travelled} was introduced to quantify the spatial movement, which was previously unspecified.  The distance feature, for example, was calculated using the Haversine formula to determine the great-circle distance between two geographic points. According to \cite{santhosh2017home}, the Haversine formula is defined as follows:
\begin{equation}
d = 2r \cdot \arcsin\sqrt{\sin^2\left(\frac{\varphi_{B} - \varphi_{A}}{2}\right) + \cos(\varphi_{A}) \cdot \cos(\varphi_{B}) \cdot \sin^2\left(\frac{\lambda_{B} - \lambda_{A}}{2}\right)},
\end{equation}
where $\varphi_{A}$ and $\varphi_{B}$ represent the latitudes, $\lambda_{A}$ and $\lambda_{B}$ represent the longitudes, $r$ represents the radius of the sphere and $d$ is the distance between the points. 
\begin{table}
    \centering
    \captionsetup{list=no} 
    \caption{Description of original and engineered features in the telemetry dataset comprising acoustic detections of dusky kob \emph{Argyrosomus japonicus} tagged and monitored in the Breede Estuary, South Africa.}
    \begin{tabular}{|p{1.5cm} p{10cm}| }
    \hline
        \multicolumn{2}{|c|}{Feature description} \\ \hline \hline
        \multicolumn{2}{|l|}{\textbf{Original Features}} \\ \hline
        \texttt{fishid} & ID code of each individual animal. \\ \hline
        \texttt{receiver} & The serial number of a specific receiver. \\ \hline
        \texttt{station} & Name of the station. \\ \hline
        \texttt{lat} & Geographic coordinate that indicates the north-south
position of a fish location. \\ \hline
        \texttt{lon} & Geographic coordinate that specifies the east-west
position of a fish location. \\ \hline
        \texttt{date} & The date on which an individual fish was detected.\\ \hline
        \texttt{time\_sa} & The time (UTC+2) at which an individual fish was detected.\\ \hline

        \multicolumn{2}{|l|}{\textbf{Engineered Features}} \\ \hline
        \texttt{distance} & The distance between two points on the Earth’s surface using the Haversine formula.\\ \hline
        \texttt{duration in same station} & The time a fish remained at the same station before moving to another station or going undetected.\\ \hline
        \texttt{num detections} &The total number of times a fish was detected during the study period.\\ \hline
        \texttt{num days detected} & The number of separate days on which a fish was detected in the Breede Estuary.\\ \hline
        \texttt{num unique stations} & The number of different stations on which a fish was detected throughout the Study period.\\ \hline
        \texttt{consecutive missing stations} & A count of the number of consecutive stations that did not detect a fish, indicating periods when the fish moved out of detection range or through areas without receiver coverage.\\ \hline
    \end{tabular}
    \label{Features_description3}
\end{table}

\subsection{Labelling process}
A detailed labelling process was first required to prepare the data for effective anomaly detection. The labelling process was based on three primary criteria, each capturing different patterns of abnormal fish movement.
\begin{itemize}
  \item \textit{If an individual fish was recorded at only one station throughout the study period, then it was flagged as an anomaly:} This criterion identifies anomalies where an individual fish was detected exclusively at a single station throughout the entire observation period (Figure \ref{fig:image4}). While dusky kob have been recorded remaining stationary for a number of days at a time \cite{naesje2012riding}, not moving at all would suggest either natural or fishing mortality, where a tagged fish has been caught and the tag discarded within the vicinity of an acoustic receiver, and as such, an anomaly. 

\begin{figure}
    \centering
    \centering
    \includegraphics[width=.8\textwidth]{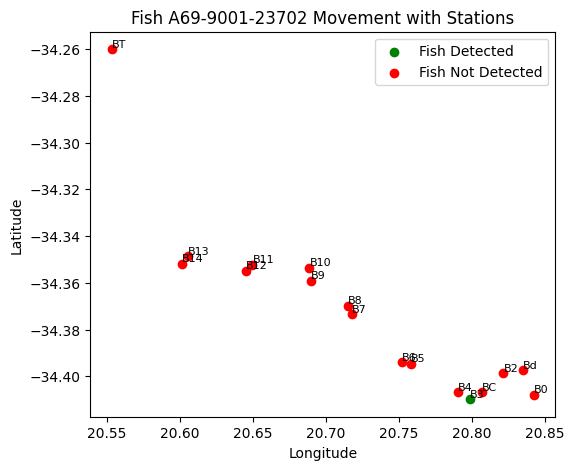} 
    \caption{Patterns of detection for dusky kob A69-9001-23702 as an example of an anomalous detection pattern where a fish was only detected at one station, suggesting it remained stationary throughout the study period.}
    \label{fig:image4}
\end{figure}

  \item \textit{If an individual moves as per normal, but then remains in the vicinity of the same station for more than 120 days:}  This criterion flags anomalies where a fish, after initially moving between stations as expected, remains at one station for more than 120 days. This prolonged stay suggests natural or fishing mortality, similar to Criterion 1.

  \item \textit{If an individual fish is not detected by consecutive stations, but is missed by more than one consecutive station:} This criterion captures anomalies where a fish is not detected at consecutive stations (missed by more than one consecutive station) (Figure \ref{fig:image3}), which could suggest code collision resulting in the creation of false of anomalous detections. Alternatively, this might suggest that the fish may have moved past the receiver at a time when reception range was reduced, preventing detection. This can occur during periods of high environmental noise caused by strong winds, currents, or high levels of boat traffic, for example.

\begin{figure}
    \centering
    \centering
    \includegraphics[width=.8\textwidth]{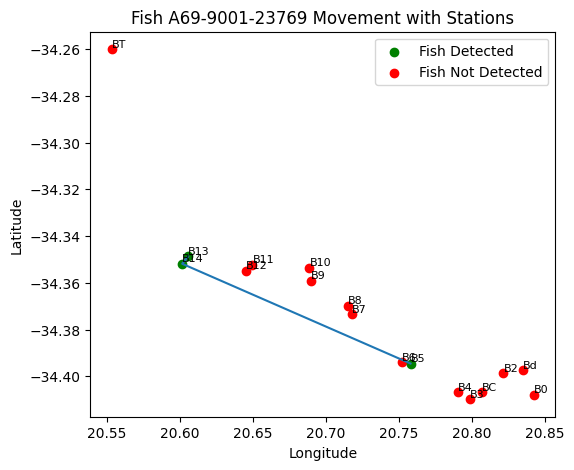} 
    \caption{An example of an anomalous detection pattern, where dusky kob A69-9001-23769 was only detected at three out of 16 stations in the estuary, but not in a consecutive manner. This pattern deviates from the defined notion of normality, as the fish was missed by multiple consecutive stations, flagging it as an anomaly.}
    \label{fig:image3}
\end{figure}
\end{itemize}

Any detection meeting at least one of the defined criteria mentioned above was marked as an anomaly (label 0), while the rest were considered normal (label 1). To ensure data quality and a robust foundation for an anomaly detection system, the labelled data underwent validation, and detected anomalies were manually reviewed by a domain expert to confirm their validity. The complete flowchart of labelling is visually depicted in Figure \ref{newFeature}.

\textcolor{black}{The initial dataset contained 3038671 detections. During data pre-processing, duplicates were removed, resulting in 3013930 unique detections. After labelling, 2754230 detections (91.38\%) were identified as \say{normal}, while 259700 (8.62\%) were classified as \say{anomalies}. The significant imbalance of 91.38\% normal detections to 8.62\% anomalies led us to choose unsupervised learning classifiers, which do not require balanced datasets, unlike supervised classifiers. The unsupervised classifiers were trained to recognize only normal patterns. When tested with both normal and anomalous patterns, they failed to recognize the anomalies, allowing us to flag such detections. Although unsupervised classifiers typically require less manual data preparation, in this study, we labelled the data to specifically train the models on normal patterns and then used the anomalous patterns to interpret the classifiers outcomes.}

\begin{figure}
    \centering
    \includegraphics[width=.8\textwidth]{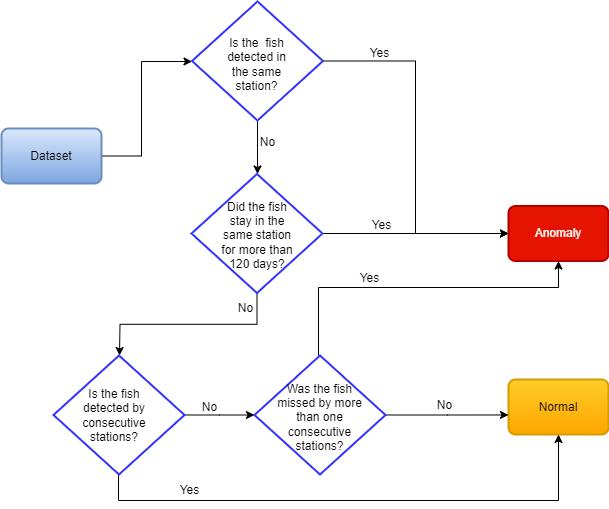}
    \caption{{\textcolor{black}{Flowchart for anomaly detection in the acoustic telemetry test dataset comprising 50 dusky kob tagged and monitored in the Breede Estuary between 2016 and 2021.}}}
    \label{newFeature}
\end{figure}

\subsection{Resampling acoustic telemetry data from real data}
\label{resampling}
\textcolor{black}{The sampling rate is determined by the frequency of normal and anomalous detections at specific times. In this study, the acoustic telemetry data were sampled irregularly on a daily basis and varied each day over the observation period. Figure \ref{WithoutResampling2} illustrates the detection of the same fish over three days. The horizontal axis represents the time of day when the fish was detected, clearly indicating that detections do not occur at regular intervals, which was expected in an acoustic telemetry dataset due to fish varying their movements on a daily basis linked to environmental variables (e.g. temperature, salinity, turbidity, river flow, etc.) and cyclical rhythms (i.e. movements influenced by time of day, tidal cycle, moon phase, etc.). This irregularity can impact the accuracy of the classifiers being trained. Since we are training unsupervised classifiers only on the normal patterns, we employ a resampling strategy that adjusts the irregular sampling of the normal samples. This strategy aims to approximate the Nyquist sampling requirements, not necessarily to balance the data. A consistent sampling rate was vital for preserving the signal's integrity, crucial for precision analysis. Well-sampled data allows models to learn more effectively, providing a reliable representation of the underlying signal, which in turn facilitates feature extraction and enhances model performance.}
\begin{figure}
    \centering
      \begin{subfigure}[t]{\textwidth}
        \centering
    \includegraphics[width=\textwidth]{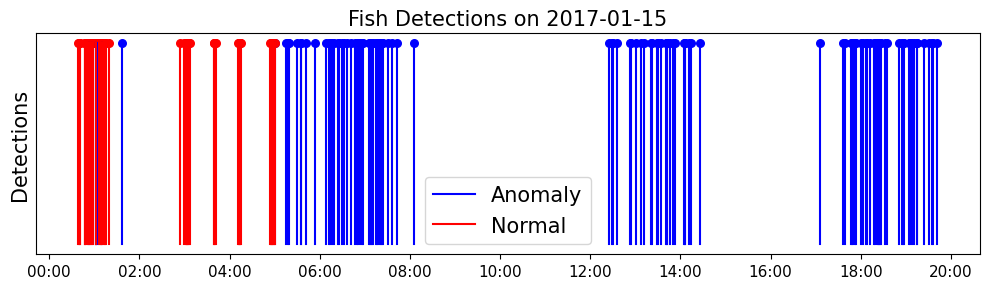}\\
           \caption{}
    \end{subfigure}\\
    \begin{subfigure}[t]{\textwidth}
        \centering\includegraphics[width=\textwidth]{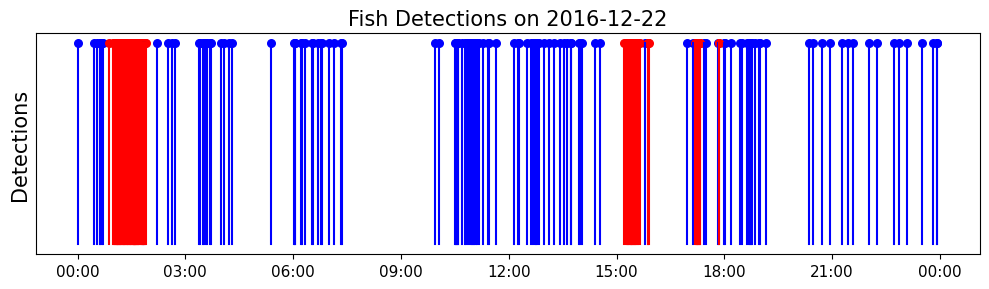}\\
               \caption{}
    \end{subfigure}\\
    \begin{subfigure}[t]{\textwidth}
        \centering\includegraphics[width=\textwidth]{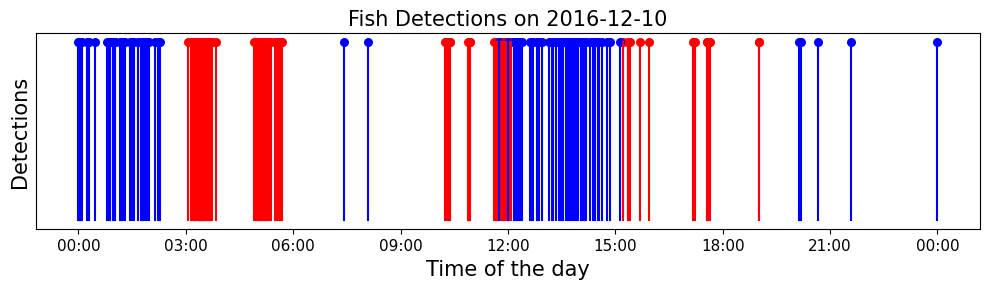}
               \caption{}
    \end{subfigure}\\
    \caption{Examples of fish detections across three different days 2017-01-15 (a), (2016-12-22 (b) and 2016-12-10(c)) show irregular sampling patterns in acoustic telemetry data. The horizontal axis represents the time of day, while the vertical markers differentiate between normal (red) and anomalous (blue) detections. The varying frequency and irregular intervals of detections show the challenges of missing data and the importance of resampling strategies to preserve signal integrity and improve unsupervised classifier performance.}
    \label{WithoutResampling2}
\end{figure}

Our strategy actively adjusted the sampling rate to reflect daily changes in data collection. This adjustment was based on calculating the minimum of the shortest non-zero time intervals between consecutive data points, labelled as {\textcolor{black}{ \(\Delta^d t\), where the subscripts $d$ indicated that this varied per day:}
\begin{equation}
 \Delta^d t = \min(\{t_{i+1} - t_i\}), ~~i=1,\cdots, N_t \label{eq:delad}
\end{equation}
{\textcolor{black}{where \( t_i \) and \( t_{i+1} \) represent the 
consecutive timestamps of data points during day $d$ and $N_t$ was the number of normal detection timestamps. To standardize the sampling rate across all days, we chose the minimum $\Delta^d t$ across days:
\begin{equation}
    \Delta t =  \min(\{\Delta^d t\}), ~~d=1,\cdots, N
\end{equation}
where $N$ was the total number of monitored days. 
The sampling rate was then the inverse of \(\Delta t\)  given as:}
\begin{equation}
   f_s = \frac{1}{\Delta t}. \label{eq:dela3}
\end{equation}
This sampling rate sets the frequency at which data points were collected, helping to generate new data points at approximately consistent intervals. 
 
After implementing the resampling strategy described in Equation \ref{eq:dela3}, we observed a pattern that approximated regular sampling. However, this was accompanied by computational challenges due to the high resampling rate, which necessitated significant computational resources to train the classifiers. Instead of using the minimum rate as specified in Equation \ref{eq:dela3}, we iteratively resampled using the next smallest rate until we found a trade-off between sampling regularity and computational feasibility for training the classifiers. With the adjustment following the trade-off search, the data in Figure \ref{WithoutResampling2} now exhibit an approximate regularity as shown in  Figure \ref{fig:same2}.}

\begin{figure}
    \centering
    \begin{subfigure}[t]{\textwidth}
    \includegraphics[width=\textwidth]{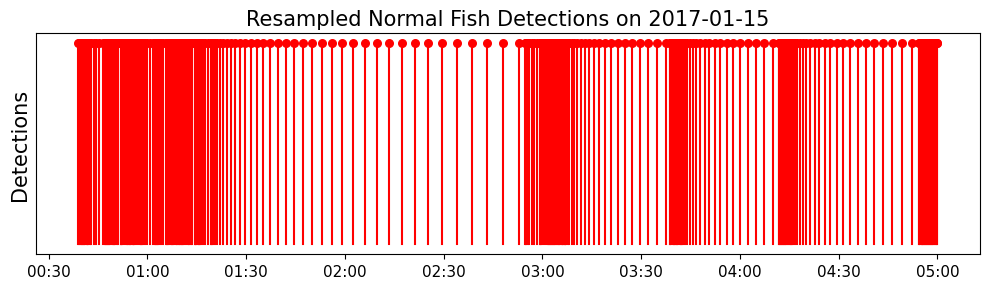}\\
                   \caption{}
    \end{subfigure}\\
    \begin{subfigure}[t]{\textwidth}
    \includegraphics[width=\textwidth]{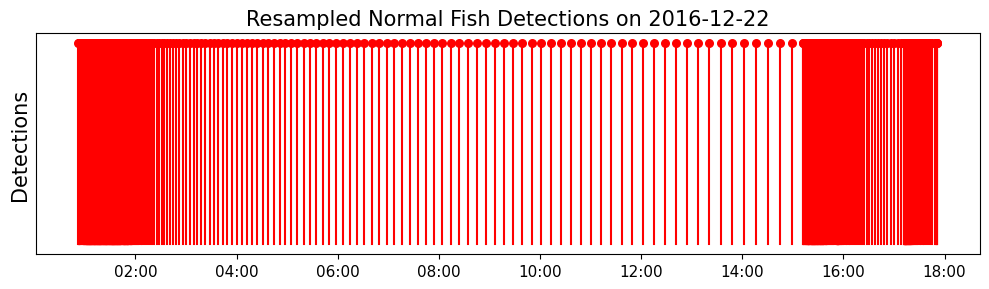}\\
                   \caption{}
    \end{subfigure}\\
    \begin{subfigure}[t]{\textwidth}
    \includegraphics[width=\textwidth]{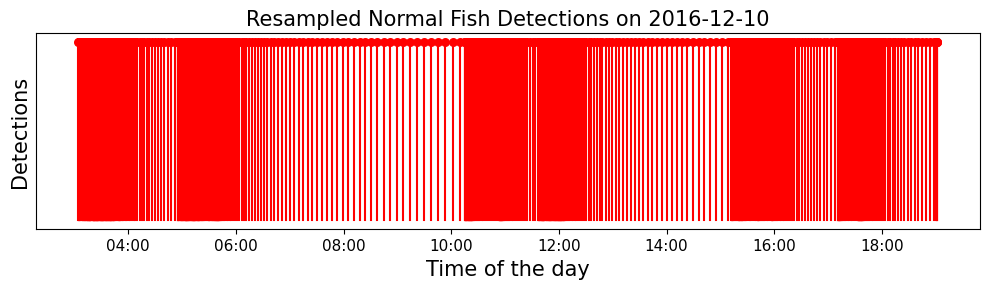}
                   \caption{}
    \end{subfigure}\\
    \caption{Comparison of resampled normal fish detections across three different days: 2017-01-15 (a), 2016-12-22 (b) and 2016-12-10 (c). Each day is resampled with the smallest sampling rate among the three days, approximating regular intervals. The resampling strategy is designed to adjust for irregular sampling patterns and ultimately improve the performance of unsupervised classifiers.} 
    \label{fig:same2}
\end{figure}
 
\subsection{Unsupervised classifiers}
\label{unsupervisedclassifiers}
{We trained three widely used traditional unsupervised classifiers for anomaly detection: IF, LOF, and DBSCAN. Additionally, we trained a NN-AE and a long short-term memory-based autoencoder (LSTM-AE).  IF employs tree structures and random partitioning to detect outliers in complex datasets. By randomly selecting features and partitioning the data to isolate observations, IF constructs isolation trees and calculates an anomaly score based on the path length from the root node to each data point — shorter paths indicate potential anomalies \cite{liu2008isolation,hariri2019extended}. LOF assesses the local density deviation of a data point relative to its neighbours, focusing on localized density rather than a global density metric. By comparing a point's density to its surrounding neighbourhood, LOF assigns an outlier score that highlights points that deviate locally, which makes it particularly suited for detecting outliers in dense data regions \cite{breunig2000lof,alghushairy2020review,cheng2019outlier}. DBSCAN is used as a density-based clustering algorithm that groups points according to density and labels any points not associated with clusters as noise or outliers. DBSCAN identifies dense regions by core points and forms clusters based on reachability; any points unreachable by core points are classified as noise, making DBSCAN highly effective for data with varied cluster shapes and densities \cite{ester1996density,khan2014dbscan,schubert2017dbscan}.} 

{Neural networks are computational models inspired by the human brain, consisting of layers of interconnected nodes (neurons) that process and learn from data \cite{nwadiugwu2020neural}. An autoencoder consists of an encoder, a latent space, and a decoder. The encoder compresses the input into a latent-space representation, and the decoder reconstructs the input data from this representation assuming a reconstruction error \cite{hinton2006reducing,hinton2006reducing,chen2018autoencoder,sakurada2014anomaly}.  The reconstruction error, or the difference between the original and reconstructed data, serves as the primary indicator of anomalies \cite{chen2018autoencoder}.  One of the challenging tasks for AEs is determining the optimal threshold to bound the reconstruction errors. In this paper, we introduced an additional block that finds this optimal threshold for the reconstruction error, and any data point exceeding this threshold was considered anomalous.}

{Although the traditional models allow quick training without extensive hyperparameter tuning, they were prone to high false normal rates (that is, they predicted anomalous as normal). This limitation arose because certain legitimate anomalies shared features with normal data points, making them difficult for traditional models to distinguish, leading to misclassifications. However, traditional models demonstrated competitive performance compared to NN-AE  in minimizing false anomalies (i.e., predicting normal instances as anomalous). The LSTM-AE quickly overfitted the data, even after resampling efforts. Due to the limitations observed with the LSTM-AE model, this paper focuses exclusively on the performance of traditional models and NN-AE as baseline models for the detection of telemetry anomalies.}

\subsection{Performance metrics}
We evaluated the models using several performance metrics, including the confusion matrix, Receiver Operating Characteristic (ROC) curve, accuracy, precision, recall, specificity, and $\text{F}_1$-score. The confusion matrix summarizes the model’s predictions into four categories: True anomalies (TA) refer to the number of anomalous samples that are correctly identified as anomalous. False anomalies (FA) are the number of normal samples incorrectly classified as anomalous. True normals (TN) represent the number of normal samples that are correctly classified as normal, while false normals (FN) are the number of anomalous samples that are incorrectly classified as normal. These values are the basis for calculating the other metrics in Table ~\ref{Classification_Measures}. The ROC curve is a graphical representation of a model’s ability to distinguish between classes. It plots the true anomaly rate against the false anomaly rate at various threshold settings. The area under the ROC curve (AUC) indicates the model’s overall performance, with a value of $1$ representing a perfect model.}
\begin{table}
\centering
\caption{Summary of key classification metrics used to evaluate model performance. Each metric provides distinct insights into the model's ability to correctly classify normal and anomalous instances.}
\begin{tabular}{p{2.5cm}p{6cm}p{4cm}}
\hline
\hline
\textbf{Name} & \textbf{Description} & \textbf{Formula} \\ 
\textbf{accuracy} & 
The proportion of correctly classified instances among the total instances.  & 
$\frac{TA + TN}{TA + TN + FA + FN}$ \\

\textbf{precision} & 
The ratio of true anomaly predictions to the total number of positive predictions, indicating how often the model is correct when it predicts a positive outcome.  & 
$\frac{TA}{TA + FA}$ \\

\textbf{recall} & 
The ratio of true anomalies to the total actual anomalies, measuring the model's ability to identify positive instances.  & 
$\frac{TA}{TA + FN}$ \\

\textbf{specificity} & 
 The ratio of true normals to the total actual negatives, indicating how well the model identifies negative instances. & 
$\frac{TN}{TN + FA}$ \\

\textbf{$\text{F}_1$-score} & 
The harmonic mean of precision and recall. It balances the two metrics, especially in cases of imbalanced class distributions. & 
$\frac{2 \times\text{precision} \times \text{recall}}{\text{precision} + \text{recall}}$ \\
\hline
\hline
\end{tabular}
\label{Classification_Measures}
\end{table}

\subsection{Experimental design}\label{experiement}
{Figure \ref{other split} provides a comprehensive overview of the experimental design, encompassing each stage from data pre-processing and dataset splitting to resampling, training, hyperparameter tuning, and testing.}

{In \textit{Step 1}, the data underwent pre-processing, which included handling missing values, removing duplicates, and performing feature engineering. After pre-processing, the data moved to \textit{Step 2} for various splits. Initially, the dataset was divided into two main categories: normal and anomalous samples. Normal samples accounted for 91.2\% of the data (2749960 samples), while anomalous samples made up 8.8\% (263970 samples) of the original dataset.}

{For the normal samples, 10\% (274996 samples) was set aside for testing, and the remaining 90\% (2474964 samples) underwent resampling as described in Section \ref{resampling}. The resampled data was then used to train the traditional models without any further splitting. For the NN-AE model, the resampled data was further divided into 80\% for training and 20\% for validation. Note that the 10\% test data was not resampled.} 

{For the anomalous samples, 50\% (131985 samples) was allocated for testing, while the other 50\% ({131985 samples) was used for training the traditional models and also served for validation to determine the autoencoder threshold, as discussed in Section \ref{unsupervisedclassifiers}.} {The splitting strategy outlined indicated that the traditional models were trained using both normal and anomalous data, while the NN-AE model was trained exclusively on normal data.}

{After \textit{Step 2}, the data proceeded to \textit{Step 3} for training and hyperparameter tuning. For each model, hyperparameter tuning was performed using the GridSearch algorithm \cite{zahedi2021search}. The IF model optimally detected anomalies by recursively partitioning data points with 100 estimators and a contamination setting of 0.001, balancing robustness and computational efficiency. The LOF identified anomalies by evaluating the local density deviation of each point relative to its neighbours, using 5 neighbours and a contamination parameter of 0.01. For DBSCAN, the optimal parameters were determined to be an epsilon value of 0.5 and a minimum sample size of 10.} {For the NN-AE, hyperparameter tuning was conducted using the validation set, which contained both normal and anomalous samples, in order to determine the optimal threshold for the autoencoder. The optimal configuration for the NN-AE consisted of a learning rate of 0.001, 128 units per layer, and 2 units in the compression layer. The model was trained with a batch size of 512 over 50 epochs. Table \ref{Hyperparameters} summarized the range of hyperparameters and their optimal values determined via GridSearch.}

{After training and hyperparameter tuning, the models proceeded to \textit{Step 4} for testing. The test data, set aside during the splitting step (prior to resampling and training), contained both normal and anomalous instances and was used to evaluate the models' ability to accurately detect anomalies.}

    \begin{figure}
        \centering
        \begin{minipage}{\textwidth}
            \centering
            \includegraphics[width=\textwidth]{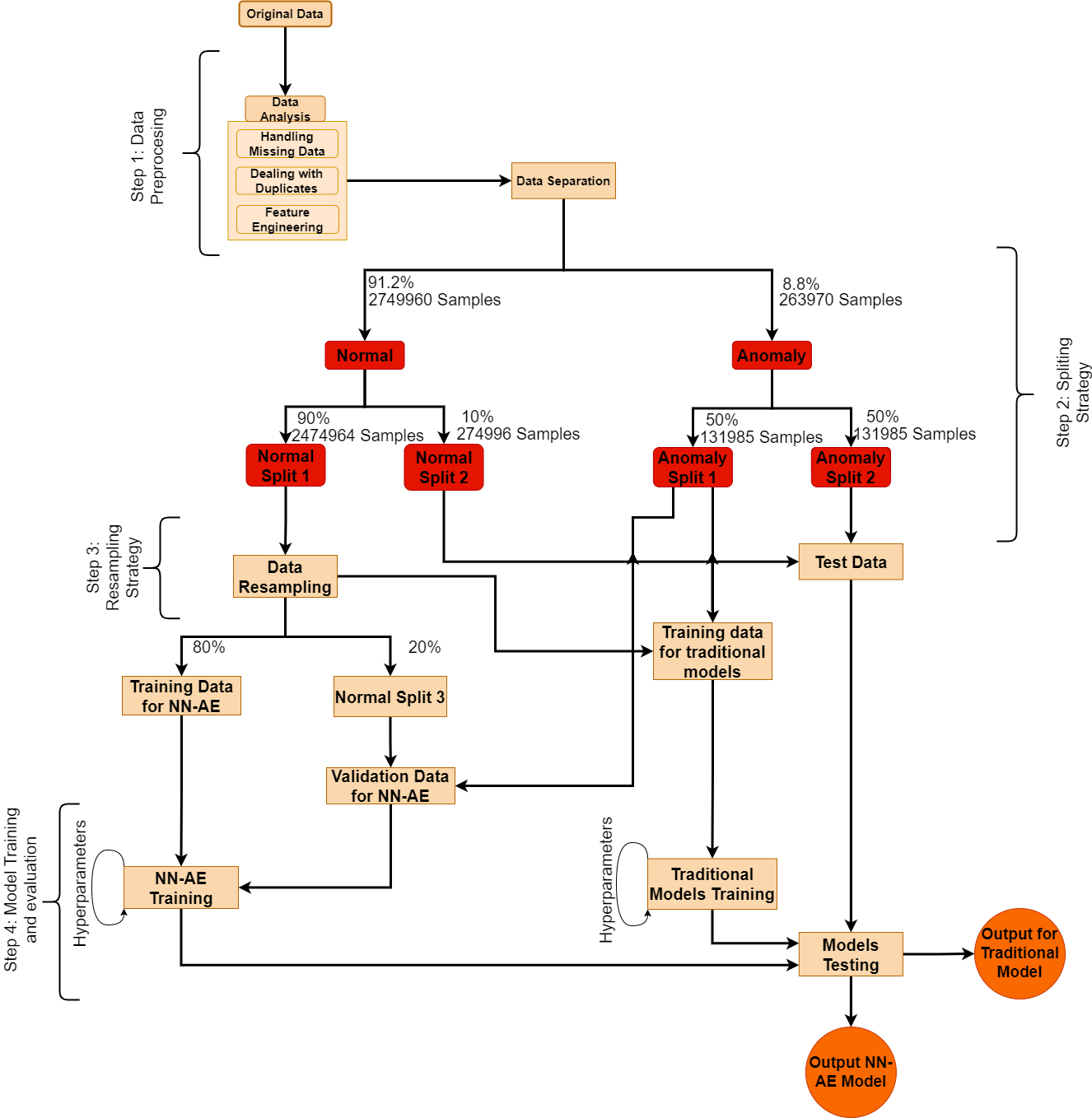}
            \caption{The diagram shows an anomaly detection workflow with data pre-processing,
followed by data splitting into training and testing sets, and concludes with
model training and testing for both the NN-AE and traditional models used in this work.}
            \label{other split}
        \end{minipage}

    \end{figure}
\section{Results and discussion}
{The results of the models, trained with the optimal hyperparameters as discussed in Section \ref{experiement}, were analysed, compared, and interpreted, and are discussed in this section.} 

\subsection{Finding optimal threshold}
\label{subsect:FOT}
Among the metrics of precision, recall, and specificity we aim to maximize the recall, as this indicates that the model has maximized the detection of true anomalies while minimizing the prediction of anomalies as normal instances (i.e., minimizing FN). This is crucial for telemetry observations, as the presence of FN in the data might skew the scientific interpretation for which the telemetry data were observed.
The goal is to achieve a recall of 1, which ensures that no FN are present in the observations. For the range of thresholds where recall is maximized, we then select the threshold that also maximizes both precision and specificity to reduce the identification of false anomalies (i.e. minimizing FA).

The algorithm works as follows. Assume that the percentiles run from $i=1,\ldots,n$ where each percentile corresponds to a specific threshold. Let $R=\{r_i\}$ be the recall set where each $r_i$ is the recall for percentile $i$, $P=\{p_j\}$ be the precision set where $j$ corresponds to selected indices from $R$, and $S=\{s_k\}$ be the specificity set where $k$ corresponds to selected indices from $P$.
The algorithm consists of three steps: the first step finds the maximum values in $R$; $R_{\mathrm{max}}=\max\{r_i: i=1,\ldots,n\}$ and identify all indices $j$ where $R_j=R_{\mathrm{max}}$, i.e. the indices of the maximum values in $R$; $J=\{j:R_j=R_{\mathrm{max}}\}$. The second step finds the maximum value in $P$ for $j\in J$; $P_{\mathrm{max}}=\max\{p_j: j\in J\}$ and identifies all indices $k$ where $P_k=P_{\mathrm{max}}$,  i.e. the indices of the maximum values in $P$; 
$K=\{k:P_k=P_{\mathrm{max}} \text{ and } k \in J\}$. The final step finds the maximum value in $S$ for $k\in K$; $S_{\mathrm{max}}=\max\{s_k: k\in K\}$ and produces all the indices $l$ where $S_l=S_{\mathrm{max}}$. The best percentile is any of the $l$ percentiles.

As shown in Tables \ref{tab:performance_metrics_1_50}, \ref{tab:performance_metrics_51_100}, and \ref{tab:performance_metrics_1_503}, the optimal percentiles identified by the algorithm are the $65^{\text{th}}$, $67^{\text{th}}$, and $69^{\text{th}}$ percentiles for the datasets without resampling, with 90s resampling, and with 65s resampling, respectively. Figure  \ref{Threshold selection} shows the corresponding threshold for $65^{\text{th}}$ percentile, where the reconstruction errors are plotted against the indices of the validation dataset without resampling. 

\begin{figure}
    \centering
    \begin{subfigure}{\textwidth}
        \centering
        \includegraphics[width=12.2cm, height=8.5cm]{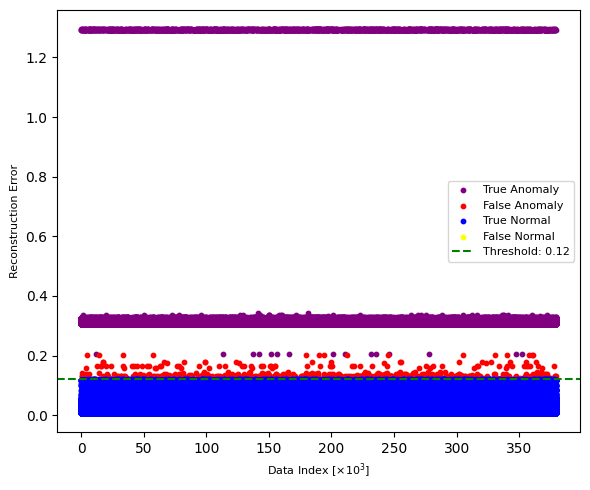}
        \label{fig:threshold_selection}
    \end{subfigure}
    
    \caption{Reconstruction errors plotted against the data index, highlighting TN (blue), TA (purple), FA (red), and FN (yellow) using the validation datasets without resampling. Higher reconstruction errors indicate potential anomalies.}
    \label{Threshold selection}
\end{figure}

\subsection{Performance}
{Using the optimal parameters, the confusion matrix generated from the classifiers' predictions on the test set is shown in Figure \ref{Matrix chart}, with the top panel depicting models trained on data without resampling, and middle and bottom panels showing models trained on resampled data at 90s and 65s, respectively. Figure \ref{fig:Rates_by_Models} provides a detailed visualization of TA, FA, TN, and FN derived from the confusion matrix of the data without resampling, with 90s resampling, and
with 65s resampling. These results offer a comprehensive overview of the models' performance, warranting a thorough discussion to evaluate their strengths and weaknesses in classifying normal and anomalous detections.}

{The results revealed that IF outperformed other models by eliminating FA, with NN-AE minimizing FA compared to LOF and DBSCAN. Resampling at finer rates decreased FA for NN-AE but not for other models, reflecting NN-AE's enhanced performance through increased data size and high-resolution data. This improvement stems from the model's ability to capture subtle patterns, reduce overfitting, and generalize more effectively with increased data size, unlike LOF and DBSCAN, which reach performance saturation quickly where additional data no longer yields significant improvement \cite{ramlakhan}.
In telemetry studies, these findings are noteworthy. 
As telemetry datasets grow continuously with each new recording, a model capable of handling large datasets is essential, and minimizing FA is important. Detecting a high rate of FA would mean that a substantial portion of normal data is misclassified as anomalous, and therefore removed from the telemetry observation. The removal of a substantial portion of normal data could skew the primary analyses for which the telemetry data are collected for. }

{The results indicate that NN-AE achieves a 100\% TA detection rate (i.e. no FN), which is crucial for telemetry observations, as it ensures that all anomalies are eliminated, leaving only normal samples. However, a very small fraction of FA; i.e. $\sim 0.00352$ of normal samples was also misclassified as anomalies and subsequently removed. This minimal removal of normal samples had no significant impact on data quality, as telemetry observations operate at a big data scale, making the loss of such a small fraction of data negligible.}  

{Figures \ref{picture} and \ref{fig:Model_Evaluation_Metrics} present the performance metrics of the models, including accuracy, $\text{F}_1$-score, recall, precision, specificity, and AUC, with error bars indicating the confidence intervals for each metric. A narrow confidence interval suggests a stable estimate, meaning repeated data reshuffling and splitting would yield similar results. In contrast, a wider confidence interval implies an unstable estimate, reducing its reliability. The results clearly demonstrate that NN-AE outperformed the other models with only specificity from other models competing comparably. This was attributed to all models showing a capability to minimize the FA, as discussed in Section \ref{subsect:FOT}. The values used to generate Figure \ref{picture} are detailed in Table \ref{tab:classifiers}, which provides a comprehensive breakdown of each metric and includes information on the computational requirements and confidence intervals of each model.}

Training times generally increase under resampling conditions. For the IF model, training times range between 4.17 minutes and 4.74 minutes. For LOF models, training times range from 24.61 minutes to 32.48 minutes. The DBSCAN model requires training times between 8.92 minutes and 9.44 minutes. All NN-AEs require significant training time, ranging from 11.6 hours to 14.2 hours. While these computation times may seem high during training, NN-AEs are generally fast when deployed. All experiments were conducted on a system with a 12th Gen Intel\textsuperscript{\textregistered} Core\texttrademark\ i7-12700 CPU @ 2.10 GHz, 16 GB RAM, running Windows 11 Enterprise (Version 23H2) with Python 3.12.7, TensorFlow 2.18.0, and scikit-learn 1.5.1.

\begin{table}
\centering
\caption{This table presents optimal hyperparameters selected for each anomaly detection model used in the work: IF, LOF, DBSCAN, and NN-AE. The models were tuned using various hyperparameter options, as listed in the "Hyperparameter" column. The "Best Parameter" columns show the selected hyperparameter values under three conditions: no resampling, resampling at 90s, and resampling at 65s.}
\begin{tabular}{lllccc}
\toprule
\textbf{Model} & \textbf{Hyperparameter} & \multicolumn{3}{c}{\textbf{Best Parameter}} \\
\cmidrule(lr){3-5}
& & \textbf{No} & \textbf{90s} & \textbf{65s} \\
\midrule
IF & Contamination: [0.001, 0.02, 0.05, 0.3] & 0.001 &0.001 & 0.001\\
 & Number of Estimators: [100, 150, 200] & 100 &100 &100 \\

\midrule
LOF & Number of Neighbors: [5, 10, 20] & 5 & 5 & 5\\
 & Contamination: [0.01, 0.08, 0.1, 0.2]] & 0.01 & 0.01 & 0.01\\

\midrule
DBSCAN & Epsilon: [ 0.5, 1.5, 2, 3.5, 5] & 0.5 & 0.5& 0.5 \\
 & Minimum Samples: [2, 4, 10] & 10 & 10& 10\\

\midrule
NN-AE & Learning Rate: [0.001, 0.01, 0.1] & 0.001 &0.001 &0.001 \\
 & Units Layer: [4, 8, 16, 32, 64, 128] & 128 & 128& 128 \\
 & Compression Layer Units: [2, 4, 8] & 2 &2 & 2\\
 & Batch Size: [128, 256, 512] & 512 &512 &512 \\
 & Epochs: [20, 50] & 50 &50 & 50\\
\bottomrule
\end{tabular}

\label{Hyperparameters}
\end{table}
\begin{figure}
\centering
\includegraphics[width=\textwidth, height=\paperheight, keepaspectratio]{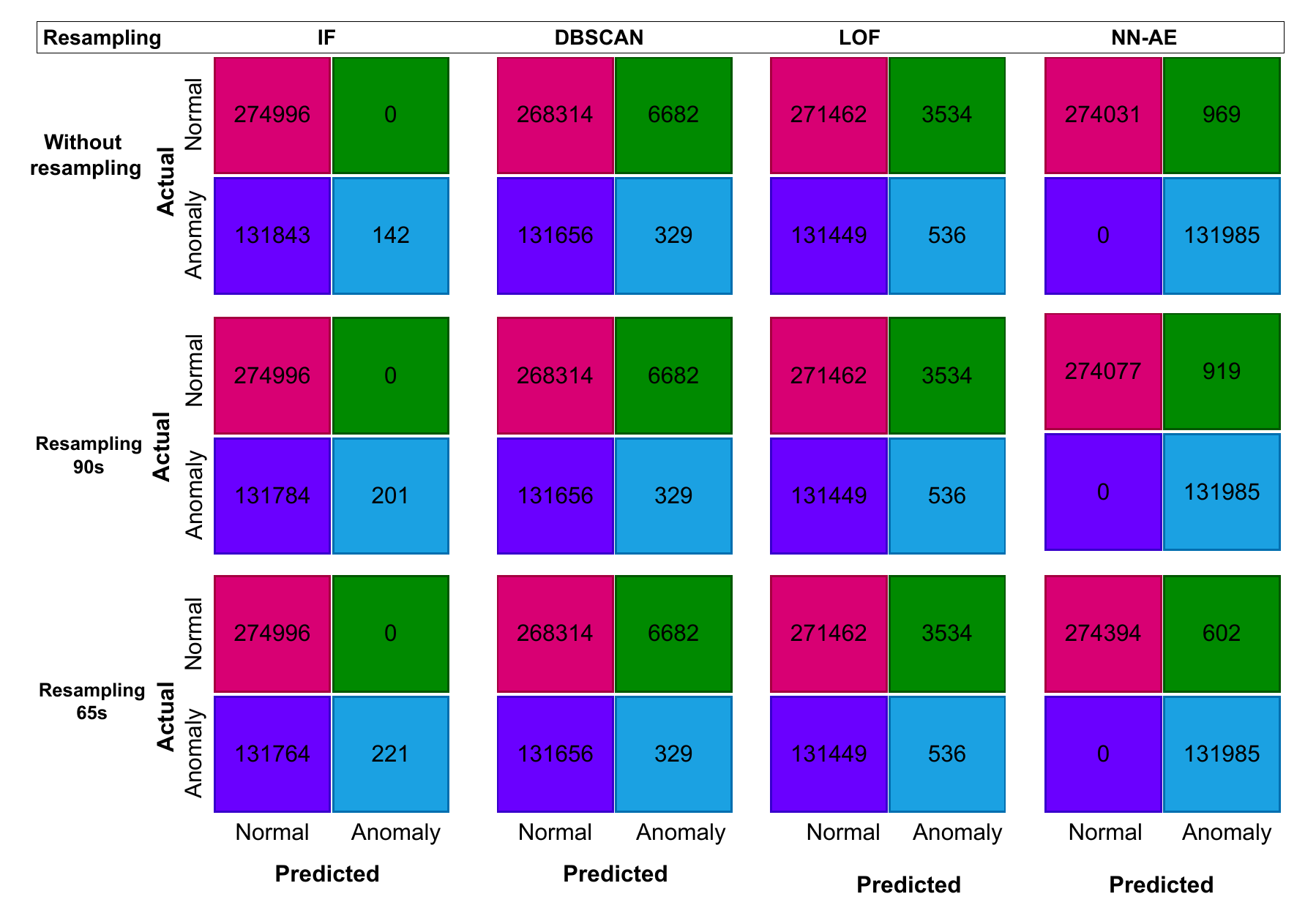}
\caption{Confusion matrices for different anomaly detection models (NN-AE, DBSCAN, IF, LOF) applied to telemetry data, comparing "Normal" and "Anomaly" classifications at different sampling rates (65s and 90s).}
\label{Matrix chart}
\end{figure}
\begin{figure}
    \centering
    \includegraphics[width=\textwidth]{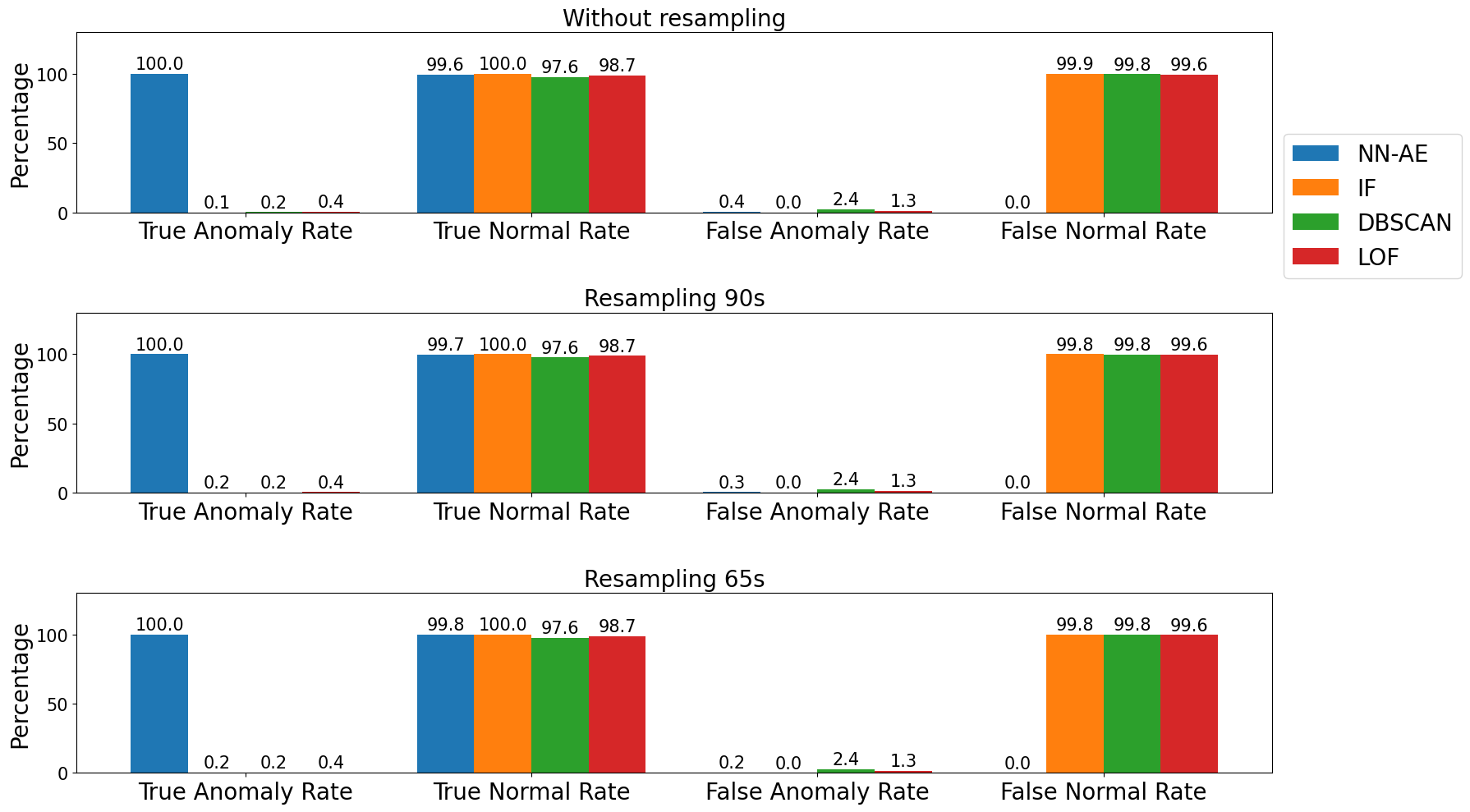}
    \caption{This bar chart shows the key metrics of the confusion matrix TA, TN, FA, and FN for anomaly detection models (NN-AE, IF, DBSCAN, LOF). The performance is compared across three conditions: without resampling, with 90s resampling, and with 65s resampling. NN-AE achieves the highest TA and TN rates, indicating its strong performance in anomaly detection. IF has a higher false normal rate, while DBSCAN and LOF have balanced but slightly lower detection capabilities.}
    \label{fig:Rates_by_Models}
\end{figure}
\begin{figure}
    \centering
    \includegraphics[width=\textwidth]{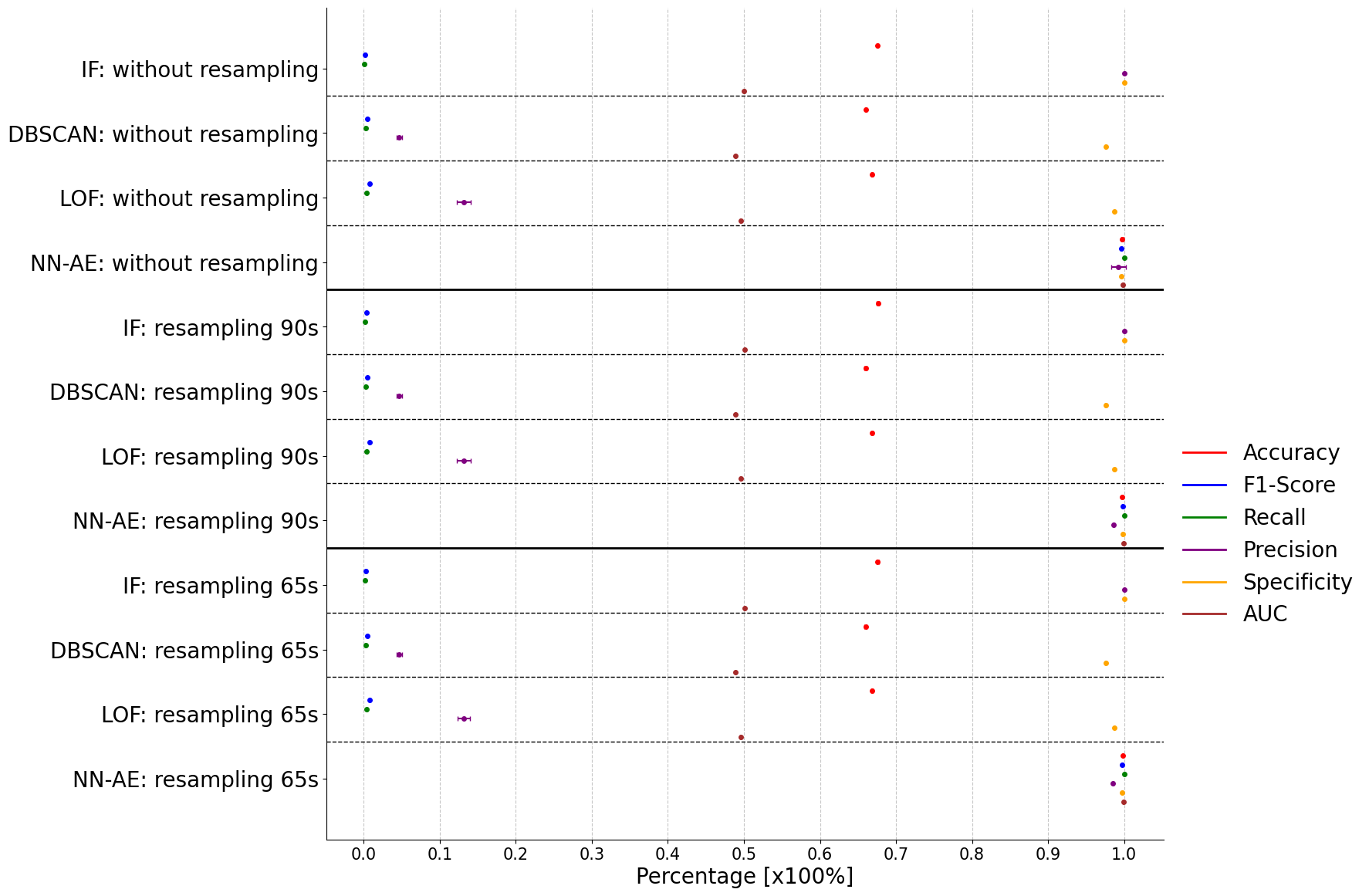}
    \caption{This figure compares the performance of different anomaly detection models (LOF, DBSCAN, IF, and NN-AE) across six metrics: accuracy, $\text{F}_1$-score, recall, precision, specificity, and AUC, along with confidence intervals. Model variations (e.g., LOF\_65s, DBSCAN\_90s) show differing levels of performance on these metrics across varying sampling rates. NN-AE performs consistently well in AUC and accuracy, while the other models exhibit more variation in precision and recall.}
    \label{picture}
\end{figure}
\subsection{Interpretability of the NN-AE detection of FA and FN}
{Understanding the mechanisms behind NN-AE models' predictions of FA and FN provides valuable insights into the nature of anomalies and their correlations with normal data. Such an analysis can reveal patterns in telemetry datasets, helping to distinguish between true anomalies and noise or artifacts in the data. This is particularly relevant for the study of telemetry, where the boundary between normal and anomalous behaviour can be complex and context-dependent. There were three types of anomalies in the dataset: (i) fish recorded at only one station, (ii) normal/expected movements, but then remained stationary for more than 120 days, and (iii) a fish passing more than one consecutively positioned receiver.} 

The NN-AE model showed strong performance in detecting anomalies in which fish were recorded at a single station throughout the study period, flagging all instances of this type of anomaly.

The second anomalous behaviour was not present in our dataset, preventing the NN-AE model from being tested on this criterion. To prepare the model for future scenarios where such behaviour might occur, synthetic data simulating such prolonged stays or expanding the dataset to capture more rare events would be beneficial. This would enhance the model’s ability to detect and learn from extreme cases, improving its overall reliability in detecting anomalies.

The model effectively identified anomalies where fish were missed by more than one consecutive station. It flagged all true anomalies without any false normals, but some normal movement patterns with occasional detection gaps were incorrectly classified as anomalies (see Appendix \ref{apendixB} for a detailed discussion). 

The models also misclassified as anomalies some normal detections that did not align with any of the three expert-defined criteria (see Appendix \ref{apendixB} for a detailed discussion), suggesting they occasionally flag unusual but not truly anomalous patterns. This over-sensitivity to movement pattern variations highlights a key challenge in automated anomaly detection for fish telemetry data: distinguishing between genuinely anomalous behaviour and natural behavioural variability.  This limitation underscores the importance of incorporating biological knowledge and expert validation in the development and refinement of machine learning models for ecological applications. Future work could focus on training models with more diverse examples of normal movement patterns or implementing additional context-aware features that better account for natural behavioural variability.

\subsection{Discussion}
{NN-AE demonstrated an exceptional performance by showing no FN, a critical factor in ensuring that no true anomalies were missed. In contrast, IF, DBSCAN, and LOF exhibited significantly higher FN fractions of $\sim 0.9989$, $\sim 0.9975$, and $\sim 0.9959$, respectively, 
on the dataset without resampling. These high FN fractions indicated that these models failed to detect nearly all true anomalies, which is a considerable limitation, particularly in applications like telemetry observations where missing anomalies can impact the interpretation of movement patterns. Moreover, these models did not show any meaningful improvement in reducing FN rates when resampling techniques were applied. This lack of responsiveness to resampling highlights their limited adaptability to variations in the data distribution and reinforces their unsuitability for high-stakes anomaly detection tasks where the identification of true anomalies is paramount. The absence of FN in NN-AE underscores its robustness and reliability in anomaly detection tasks, making it a superior choice for datasets where accuracy in identifying anomalies is essential. This advantage is particularly pronounced in dynamic and large-scale datasets, such as those encountered in telemetry studies, where ensuring the complete identification of anomalies is critical for maintaining the integrity of the analysis.}

{In terms of FA, IF showed no FA detections, while NN-AE reports a negligible FA fraction of $\sim 0.00352$. This minimal FA rate in NN-AE further decreased when resampling techniques were applied, demonstrating its adaptability and capacity to improve its performance with data resampling. On the other hand, DBSCAN and LOF exhibited FA fractions of approximately $\sim 0.0242$ and $\sim 0.0128$, respectively, on the dataset without resampling. Unlike NN-AE, these models showed no reduction in FA rates even with the application of resampling techniques, highlighting their limited ability to adapt to data resampling.}

{The differences in FN and FA rates among these models underscore their varying capabilities and limitations in anomaly detection tasks. NN-AE's superior performance, marked by the absence of FN detections and negligible FA rates that further reduce with resampling, makes it particularly suitable for critical applications like telemetry, where both the identification of TA and the minimization of FA are essential. }

{While this work demonstrates the impressive capabilities of the NN-AE model in accurately detecting anomalies, it also brings to light certain limitations that warrant further discussion. One significant challenge was the computational demand associated with training and optimizing the NN-AE model. Given the complexity of the architecture and the volume of data often encountered in telemetry and other large-scale applications, the model required substantial computational resources, which can be a limiting factor in environments with constrained infrastructure or limited access to high-performance computing. Another noteworthy limitation was the occurrence of false anomaly detections, where normal behaviour was misclassified as anomalous. Although the FA fraction $\sim 0.00352$ observed in this study was exceptionally low, such misclassifications, however negligible, may still have implications depending on the application. For instance, in critical domains like telemetry, monitoring even minimal false anomalies could lead to unnecessary interventions in the field. Nonetheless, it is worth emphasizing that the FA rate of the NN-AE model was significantly negligible compared to alternative models like DBSCAN or LOF, which exhibit much higher FA rates and failed to improve with resampling. The balance between computational efficiency and detection accuracy poses an ongoing trade-off in anomaly detection frameworks. While the NN-AE model offers a clear advantage in terms of recall, precision, and adaptability, its computational intensity highlights the need for future research into more resource-efficient algorithms that retain comparable performance. Additionally, incorporating deep learning explainability mechanisms into the NN-AE model could enhance its applicability by providing insights into the specific features or patterns leading to false anomaly detections. This could help refine the model further and reduce even the negligible rates of misclassification.}

\section{Conclusion}
{This study focused on dusky kob acoustically tagged in the Breede Estuary, providing valuable insights into anomaly detection in fish movement patterns using acoustic telemetry data. While the findings are robust, their specificity may limit their generalizability to other species, estuaries, or regions. Variations in environmental conditions, such as water temperature, salinity, or habitat structure, necessitate recalibrating or retraining models when applied to new ecosystems. Furthermore, movement patterns, habitat preferences, and/or behaviours are often species-specific, and as such, each dataset may exhibit anomalies differently, requiring tailored feature engineering and anomaly detection techniques to maintain accuracy.}

{The results clearly demonstrate the effectiveness of the NN-AE model in anomaly detection. With accuracy, precision, and recall significantly outperforming traditional unsupervised learning models such as IF, LOF, and DBSCAN across all evaluation metrics, except for specificity, where IF achieved the highest performance, followed closely by the NN-AE model. These results underscore the NN-AE model's ability to capture complex temporal dependencies in time series data, enabling the detection of subtle anomalies that traditional methods often overlook. Such capabilities are critical for telemetry studies and conservation efforts, where understanding anomalies in fish behaviour is essential for assessing ecosystem dynamics, habitat usage, and responses to environmental changes.}

{The practical implications of this work are substantial. By automating anomaly detection, the NN-AE model reduces the need for labour-intensive manual data review, enabling researchers to focus on higher-level ecological analyses and conservation strategy development. In fisheries management, early detection of behavioural anomalies could alert stakeholders to threats such as habitat degradation, poaching, or disease outbreaks, facilitating timely interventions.}

{However, this study also highlights areas for improvement and future exploration. Expanding the dataset to include a broader range of species and environmental conditions is critical to improving the generalizability and robustness of the model. Enhancing the model's ability to detect slow deviations from expected movement patterns remains a priority. Exploring adaptive thresholds that dynamically adjust based on temporal data patterns could refine detection capabilities, particularly for subtle, gradual anomalies. Furthermore, integrating attention-based approaches could further enhance its capacity to minimise false anomalies.} 

{In conclusion, this study demonstrated the transformative potential of machine learning, particularly NN-AE models, in automating anomaly detection in acoustic telemetry data. Beyond technical success, the ecological significance of this work is profound. Detecting movement anomalies provides crucial insights into environmental changes, human impacts, and fish health. This information is instrumental in developing conservation measures, such as identifying signs of habitat destruction, illegal fishing activities, or climate change effects on aquatic ecosystems. While the findings are directly applicable to dusky kob in the Breede Estuary, the methodology provides a scalable framework for studying other species and environments, contributing to global conservation efforts and advancing the role of artificial intelligence in ecological research.}

\section*{CRediT authorship contribution statement}
\textbf{Siphendulwe Zaza:} Conceptualization, Data curation, Formal analysis, Investigation, Methodology, Software, Visualisation, Writing – original draft, Writing – review \& editing. 
\textbf{Marcellin Atemkeng:} Conceptualization, Methodology, Visualisation,  Formal analysis, Resources, Writing – review \& editing, Supervision, Validation. 
\textbf{Taryn S. Murray:} Conceptualization, Formal analysis, Visualisation, Data curation, Resources, Fieldwork, Writing – review \& editing, Supervision, Validation. 
\textbf{John David Filmalter:} Data curation, Fieldwork, Resources, Writing – review \& editing, Validation. 
\textbf{Paul D. Cowley:} Funding acquisition, Resources, Validation.

\section*{Data availability}
Data will be made available on request.

\section*{Acknowledgments}
The Acoustic Tracking Array Platform (ATAP) is hosted by the South African Institute for Aquatic Biodiversity, a National Facility of the National Research Foundation (NRF-SAIAB). The Ocean Tracking Network (OTN) headquartered at Dalhousie University, Canada, the Department of Science and Innovation-Shallow Marine and Coastal Research Infrastructure program, and the South African Environmental Observation Network Elwandle Node (NRF-SAEON Elwandle Node) are thanked for providing acoustic telemetry hardware that facilitated data collection for this study. Furthermore, the Save Our Seas Foundation and the African Coelacanth Ecosystem Programme are acknowledged for funding to maintain the national ATAP. Dr JD Filmalter is thanked for his efforts in tagging all fish for this study, as well as maintaining the acoustic receiver array positioned in the Breede Estuary. Further, funding for acoustic transmitters and project running expenses were funded by the South Africa–Norway Cooperation on Ocean Research (SANOCEAN Project No. 287015). Siphendulwe Zaza thanks  LEVENSTEIN, THE ADA \& BERTIE for the financial support. Marcellin Atemkeng thanks the National Research Foundation of South Africa for support through project number CSRP23040990793.

\begin{appendices}
\section{}
Tables \ref{tab:performance_metrics_1_50}, \ref{tab:performance_metrics_1_503}, and \ref{tab:performance_metrics_1_502} show the optimal thresholds for NN-AE trained and validated on three datasets: the original dataset without resampling, with 90s resampling, and with 65s resampling, respectively. The optimal percentile thresholds were 65 for the model trained on the dataset without resampling, 67 for the 90s resampling dataset, and 69 for the 65s  resampling dataset. This trend suggests that datasets with higher temporal resolution (i.e., shorter resampling intervals) required slightly higher percentile thresholds for optimal anomaly detection.

\begin{table}
    \centering
        \caption{Performance metrics  at different percentiles of reconstruction errors, demonstrating the trade-offs, and supporting the selection of the 65th percentile as the optimal threshold for anomaly detection in data without resampling.}
    \begin{adjustbox}{width=\textwidth}
    \begin{tabular}{ccccccc}
        \toprule
        Percentile & Optimal Threshold & precision & recall & $\text{F}_1$-score & specificity & accuracy \\
        \midrule
        1  & 0.007764 & 0.351275 & 1.0 & 0.519917 & 0.015156 & 0.357688 \\
2  & 0.008336 & 0.354852 & 1.0 & 0.523824 & 0.030457 & 0.367667 \\
3  & 0.008481 & 0.358488 & 1.0 & 0.527775 & 0.045702 & 0.377609 \\
4& 0.009629 & 0.362266 & 1.0 & 0.531858 & 0.061213 & 0.387726 \\
5  & 0.009824 & 0.366179 & 1.0 & 0.536064 & 0.076946 & 0.397987 \\
6  & 0.010049 & 0.370074 & 1.0 & 0.540225 & 0.092272 & 0.407982 \\
7  & 0.011195 & 0.374071 & 1.0 & 0.544471 & 0.107670 & 0.418025 \\
8  & 0.011339 & 0.378028 & 1.0 & 0.548650 & 0.122591 & 0.427757 \\
9  & 0.011383 & 0.382179 & 1.0 & 0.553009 & 0.137913 & 0.437749 \\
10 & 0.011512 & 0.386402 & 1.0 & 0.557417 & 0.153165 & 0.447697 \\
11 & 0.011848 & 0.390943 & 1.0 & 0.562126 & 0.169194 & 0.458151 \\
12 & 0.012492 & 0.395418 & 1.0 & 0.566738 & 0.184633 & 0.468220 \\
13 & 0.012549 & 0.399941 & 1.0 & 0.571369 & 0.199885 & 0.478168 \\
14 & 0.012802 & 0.404536 & 1.0 & 0.576042 & 0.215029 & 0.488044 \\
15 & 0.013133 & 0.409309 & 1.0 & 0.580865 & 0.230403 & 0.498071 \\
16 & 0.014166 & 0.414214 & 1.0 & 0.585786 & 0.245829 & 0.508132 \\
17 & 0.014493 & 0.419164 & 1.0 & 0.590719 & 0.261033 & 0.518048 \\
18 & 0.014778 & 0.424363 & 1.0 & 0.595864 & 0.276622 & 0.528215 \\
19 & 0.015240 & 0.429626 & 1.0 & 0.601033 & 0.292016 & 0.538255 \\
20 & 0.015895 & 0.435106 & 1.0 & 0.606375 & 0.307648 & 0.548450 \\
21 & 0.016621 & 0.440711 & 1.0 & 0.611796 & 0.323236 & 0.558617 \\
22 & 0.017105 & 0.446431 & 1.0 & 0.617286 & 0.338739 & 0.568728 \\
23 & 0.018964 & 0.452512 & 1.0 & 0.623075 & 0.354792 & 0.579197 \\
24 & 0.019656 & 0.458394 & 1.0 & 0.628629 & 0.369916 & 0.589061 \\
25 & 0.020551 & 0.464440 & 1.0 & 0.634290 & 0.385059 & 0.598937 \\
26 & 0.021317 & 0.470800 & 1.0 & 0.640196 & 0.400571 & 0.609054 \\
27 & 0.021895 & 0.477062 & 1.0 & 0.645961 & 0.415439 & 0.618751 \\
28 & 0.022122 & 0.483591 & 1.0 & 0.651919 & 0.430530 & 0.628594 \\
29 & 0.023237 & 0.490313 & 1.0 & 0.658000 & 0.445650 & 0.638455 \\
30 & 0.023745 & 0.497471 & 1.0 & 0.664415 & 0.461299 & 0.648661 \\
31 & 0.024585 & 0.504572 & 1.0 & 0.670719 & 0.476386 & 0.658500 \\
32 & 0.025161 & 0.511943 & 1.0 & 0.677199 & 0.491602 & 0.668424 \\
33 & 0.025700 & 0.519722 & 1.0 & 0.683970 & 0.507194 & 0.678593 \\
34 & 0.027399 & 0.527657 & 1.0 & 0.690806 & 0.522625 & 0.688657 \\
35 & 0.029553 & 0.535921 & 1.0 & 0.697850 & 0.538209 & 0.698821 \\
36 & 0.029990 & 0.544445 & 1.0 & 0.705037 & 0.553789 & 0.708982 \\
37 & 0.030776 & 0.553183 & 1.0 & 0.712322 & 0.569259 & 0.719072 \\
38 & 0.031713 & 0.562112 & 1.0 & 0.719682 & 0.584573 & 0.729060 \\
39 & 0.032265 & 0.571391 & 1.0 & 0.727242 & 0.599979 & 0.739108 \\
40 & 0.033151 & 0.581078 & 1.0 & 0.735041 & 0.615539 & 0.749256 \\
41 & 0.034136 & 0.590770 & 1.0 & 0.742747 & 0.630594 & 0.759074 \\
42 & 0.035481 & 0.600762 & 1.0 & 0.750595 & 0.645608 & 0.768867 \\
43 & 0.037271 & 0.611319 & 1.0 & 0.758781 & 0.660937 & 0.778864 \\
44 & 0.038495 & 0.622107 & 1.0 & 0.767036 & 0.676065 & 0.788730 \\
45 & 0.038779 & 0.633444 & 1.0 & 0.775593 & 0.691406 & 0.798736 \\
46 & 0.039302 & 0.644767 & 1.0 & 0.784022 & 0.706190 & 0.808378 \\
47& 0.041025 & 0.656769 & 1.0 & 0.792831 & 0.721306 & 0.818236 \\
48 & 0.041799 & 0.669638 & 1.0 & 0.802136 & 0.736910 & 0.828413 \\
49 & 0.043906 & 0.682682 & 1.0 & 0.811421 & 0.752126 & 0.838338 \\
50 & 0.044223 & 0.696193 & 1.0 & 0.820889 & 0.767286 & 0.848225 \\
        \bottomrule
    \end{tabular}
    \end{adjustbox}
    \label{tab:performance_metrics_1_50}
\end{table}
\addtocounter{table}{-1}
\begin{table}
\renewcommand\thetable{\ref{tab:performance_metrics_1_50}}
    \centering
    \caption{Cont.}
    \begin{adjustbox}{width=\textwidth}
    \begin{tabular}{ccccccc}
        \toprule
        Percentile & Optimal Threshold & precision & recall & $\text{F}_1$-score & specificity & accuracy \\
        \midrule
 51 & 0.044478 & 0.710544 & 1.0 & 0.830781 & 0.782757 & 0.858315 \\
52 & 0.046398 & 0.725117 & 1.0 & 0.840658 & 0.797840 & 0.868152 \\
53 & 0.049362 & 0.740752 & 1.0 & 0.851071 & 0.813363 & 0.878276 \\
54 & 0.052883 & 0.756430 & 1.0 & 0.861327 & 0.828285 & 0.888008 \\
55 & 0.053363 & 0.773131 & 1.0 & 0.872052 & 0.843513 & 0.897940 \\
56 & 0.056369 & 0.790504 & 1.0 & 0.882996 & 0.858673 & 0.907827 \\
57 & 0.060077 & 0.809227 & 1.0 & 0.894556 & 0.874281 & 0.918007 \\
58 & 0.066410 & 0.828651 & 1.0 & 0.906297 & 0.889728 & 0.928081 \\
59 & 0.081708 & 0.849439 & 1.0 & 0.918591 & 0.905478 & 0.938353 \\
60 & 0.084467 & 0.869931 & 1.0 & 0.930442 & 0.920266 & 0.947998 \\
61 & 0.098710 & 0.892116 & 1.0 & 0.942982 & 0.935510 & 0.957940 \\
62 & 0.101801 & 0.915615 & 1.0 & 0.955949 & 0.950852 & 0.967946 \\
63 & 0.117203 & 0.940660 & 1.0 & 0.969423 & 0.966359 & 0.978060 \\
64 & 0.117517 & 0.966909 & 1.0 & 0.983176 & 0.981749 & 0.988097 \\
\textcolor{red}{65} & \textcolor{red}{0.120692} & \textcolor{red}{0.993855} & \textcolor{red}{1.0} & \textcolor{red}{0.996918} & \textcolor{red}{0.996703} & \textcolor{red}{0.997850} \\
66 & 0.305324 & 1.0 & 0.977596 & 0.988671 & 1.0 & 0.992208 \\
67 & 0.305671 & 1.0 & 0.948721 & 0.973686 & 1.0 & 0.982165 \\
68 & 0.305921 & 1.0 & 0.920021 & 0.958345 & 1.0 & 0.972183 \\
69 & 0.306270 & 1.0 & 0.891253 & 0.942500 & 1.0 & 0.962177 \\
70 & 0.306705 & 1.0 & 0.862537 & 0.926196 & 1.0 & 0.952190 \\
71 & 0.307026 & 1.0 & 0.833867 & 0.909409 & 1.0 & 0.942219 \\
72 & 0.307326 & 1.0 & 0.805069 & 0.892009 & 1.0 & 0.932202 \\
73 & 0.307718 & 1.0 & 0.776369 & 0.874108 & 1.0 & 0.922220 \\
74 & 0.308059 & 1.0 & 0.747593 & 0.855568 & 1.0 & 0.912212 \\
75 & 0.308449 & 1.0 & 0.718748 & 0.836363 & 1.0 & 0.902180 \\
76 & 0.308957 & 1.0 & 0.689980 & 0.816554 & 1.0 & 0.892174 \\
77 & 0.310452 & 1.0 & 0.661234 & 0.796076 & 1.0 & 0.882176 \\
78 & 0.312662 & 1.0 & 0.632534 & 0.774911 & 1.0 & 0.872194 \\
79 & 0.313046 & 1.0 & 0.603788 & 0.752953 & 1.0 & 0.862196 \\
80 & 0.313664 & 1.0 & 0.575050 & 0.730199 & 1.0 & 0.852201 \\
81 & 0.316172 & 1.0 & 0.546297 & 0.706587 & 1.0 & 0.842201 \\
82 & 0.317073 & 1.0 & 0.517521 & 0.682061 & 1.0 & 0.832192 \\
83 & 0.320032 & 1.0 & 0.488798 & 0.656634 & 1.0 & 0.822202 \\
84 & 0.320922 & 1.0 & 0.466833 & 0.636519 & 1.0 & 0.814563 \\
85 & 0.320926 & 1.0 & 0.432193 & 0.603540 & 1.0 & 0.802515 \\
86 & 0.320967 & 1.0 & 0.402402 & 0.573875 & 1.0 & 0.792154 \\
87 & 0.321086 & 1.0 & 0.373489 & 0.543855 & 1.0 & 0.782098 \\
88 & 0.321163 & 1.0 & 0.345032 & 0.513046 & 1.0 & 0.772200 \\
89 & 0.321251 & 1.0 & 0.316657 & 0.481002 & 1.0 & 0.762331 \\
90 & 0.321349 & 1.0 & 0.287464 & 0.446559 & 1.0 & 0.752178 \\
91 & 0.321493 & 1.0 & 0.258787 & 0.411169 & 1.0 & 0.742204 \\
92 & 0.322034 & 1.0 & 0.230026 & 0.374018 & 1.0 & 0.732201 \\
93 & 0.323296 & 1.0 & 0.201265 & 0.335089 & 1.0 & 0.722198 \\
94 & 0.324078 & 1.0 & 0.172535 & 0.294294 & 1.0 & 0.712205 \\
95 & 0.325454 & 1.0 & 0.143721 & 0.251322 & 1.0 & 0.702184 \\
96 & 0.325955 & 1.0 & 0.115059 & 0.206372 & 1.0 & 0.692215 \\
97 & 0.326316 & 1.0 & 0.086260 & 0.158820 & 1.0 & 0.682198 \\
98 & 0.326627 & 1.0 & 0.057567 & 0.108867 & 1.0 & 0.672219 \\
99 & 0.326864 & 1.0 & 0.028738 & 0.055871 & 1.0 & 0.662192 \\
100 & 1.294590 & 0.0 & 0.0 & 0.0 & 1.0 & 0.652197 \\
        \bottomrule
    \end{tabular}
    \end{adjustbox}
    \label{tab:performance_metrics_51_100}
\end{table}

\begin{table}
    \centering
        \caption{Performance metrics at different
percentiles of reconstruction errors, demonstrating the trade-offs, and supporting the
selection of the 67th percentile as the optimal threshold for anomaly detection with   90s resampling data.}
    \begin{adjustbox}{width=\textwidth}
    \begin{tabular}{ccccccc}
        \toprule
        Percentile & Optimal Threshold & precision & recall & $\text{F}_1$-score & specificity & accuracy \\
        \midrule
1 & 0.006183 & 0.328163 & 1.0 & 0.494161 & 0.014839 & 0.334894 \\
2 & 0.007279 & 0.331551 & 1.0 & 0.497993 & 0.029824 & 0.345010 \\
3 & 0.008931 & 0.334971 & 1.0 & 0.501840 & 0.044641 & 0.355014 \\
4 & 0.009157 & 0.338433 & 1.0 & 0.505715 & 0.059334 & 0.364934 \\
5 & 0.009793 & 0.342001 & 1.0 & 0.509688 & 0.074169 & 0.374949 \\
6 & 0.010017 & 0.345617 & 1.0 & 0.513693 & 0.088892 & 0.384889 \\
7 & 0.010216 & 0.349458 & 1.0 & 0.517923 & 0.104194 & 0.395219 \\
8 & 0.010680 & 0.353281 & 1.0 & 0.522110 & 0.119094 & 0.405279 \\
9 & 0.011087 & 0.357282 & 1.0 & 0.526467 & 0.134349 & 0.415578 \\
10 & 0.011549 & 0.361237 & 1.0 & 0.530748 & 0.149093 & 0.425532 \\
11 & 0.011866 & 0.365266 & 1.0 & 0.535084 & 0.163790 & 0.435454 \\
12 & 0.011926 & 0.369464 & 1.0 & 0.539574 & 0.178757 & 0.445559 \\
13 & 0.012397 & 0.373673 & 1.0 & 0.544049 & 0.193428 & 0.455464 \\
14 & 0.012918 & 0.377979 & 1.0 & 0.548599 & 0.208099 & 0.465368 \\
15 & 0.013253 & 0.382510 & 1.0 & 0.553356 & 0.223179 & 0.475549 \\
16 & 0.013569 & 0.386962 & 1.0 & 0.557999 & 0.237653 & 0.485321 \\
17 & 0.014060 & 0.391588 & 1.0 & 0.562793 & 0.252343 & 0.495238 \\
18 & 0.014642 & 0.396336 & 1.0 & 0.567680 & 0.267065 & 0.505178 \\
19 & 0.015152 & 0.401312 & 1.0 & 0.572766 & 0.282119 & 0.515341 \\
20 & 0.015383 & 0.406249 & 1.0 & 0.577777 & 0.296692 & 0.525180 \\
21 & 0.016471 & 0.411359 & 1.0 & 0.582926 & 0.311407 & 0.535114 \\
22 & 0.016972 & 0.416781 & 1.0 & 0.588349 & 0.326625 & 0.545388 \\
23 & 0.017516 & 0.422152 & 1.0 & 0.593681 & 0.341314 & 0.555305 \\
24 & 0.018733 & 0.427626 & 1.0 & 0.599073 & 0.355905 & 0.565156 \\
25 & 0.019276 & 0.433480 & 1.0 & 0.604794 & 0.371102 & 0.575415 \\
26 & 0.019818 & 0.439228 & 1.0 & 0.610366 & 0.385631 & 0.585224 \\
27 & 0.020752 & 0.445244 & 1.0 & 0.616151 & 0.400433 & 0.595218 \\
28 & 0.021397 & 0.451311 & 1.0 & 0.621936 & 0.414962 & 0.605027 \\
29 & 0.021789 & 0.457673 & 1.0 & 0.627950 & 0.429783 & 0.615033 \\
30 & 0.022994 & 0.464117 & 1.0 & 0.633989 & 0.444381 & 0.624888 \\
31 & 0.023402 & 0.470948 & 1.0 & 0.640332 & 0.459421 & 0.635042 \\
32 & 0.023835 & 0.477815 & 1.0 & 0.646651 & 0.474107 & 0.644957 \\
33 & 0.024215 & 0.484927 & 1.0 & 0.653132 & 0.488876 & 0.654928 \\
34 & 0.024861 & 0.492325 & 1.0 & 0.659810 & 0.503788 & 0.664995 \\
35 & 0.025308 & 0.499811 & 1.0 & 0.666498 & 0.518427 & 0.674878 \\
36 & 0.026076 & 0.507678 & 1.0 & 0.673456 & 0.533346 & 0.684950 \\
37 & 0.027332 & 0.515961 & 1.0 & 0.680705 & 0.548564 & 0.695225 \\
38 & 0.027971 & 0.524266 & 1.0 & 0.687893 & 0.563337 & 0.705198 \\
39 & 0.028229 & 0.532976 & 1.0 & 0.695348 & 0.578337 & 0.715325 \\
40 & 0.028868 & 0.541850 & 1.0 & 0.702857 & 0.593124 & 0.725308 \\
41 & 0.030642 & 0.551010 & 1.0 & 0.710517 & 0.607887 & 0.735275 \\
42 & 0.032440 & 0.560660 & 1.0 & 0.718491 & 0.622919 & 0.745424 \\
43 & 0.033150 & 0.570766 & 1.0 & 0.726736 & 0.638115 & 0.755683 \\
44 & 0.034302 & 0.580823 & 1.0 & 0.734836 & 0.652714 & 0.765539 \\
45 & 0.035420 & 0.591328 & 1.0 & 0.743188 & 0.667432 & 0.775475 \\
46 & 0.036268 & 0.602303 & 1.0 & 0.751796 & 0.682260 & 0.785486 \\
47 & 0.037056 & 0.613504 & 1.0 & 0.760462 & 0.696848 & 0.795335 \\
48 & 0.037971 & 0.625625 & 1.0 & 0.769704 & 0.712044 & 0.805594 \\
49 & 0.040528 & 0.638043 & 1.0 & 0.779031 & 0.727014 & 0.815701 \\
50 & 0.043400 & 0.650496 & 1.0 & 0.788243 & 0.741452 & 0.825448 \\
        \bottomrule
    \end{tabular}
    \end{adjustbox}
    \label{tab:performance_metrics_1_503}
\end{table}

\begin{table}
\renewcommand\thetable{\ref{tab:performance_metrics_1_503}}
    \centering
    \caption{Cont.}
    \begin{adjustbox}{width=\textwidth}
    \begin{tabular}{ccccccc}
        \toprule
        Percentile & Optimal Threshold & precision & recall & $\text{F}_1$-score & specificity & accuracy \\
        \midrule
51 & 0.045826 & 0.663835 & 1.0 & 0.797958 & 0.756317 & 0.835483 \\
52 & 0.048199 & 0.677659 & 1.0 & 0.807863 & 0.771104 & 0.845467 \\
53 & 0.050456 & 0.692141 & 1.0 & 0.818065 & 0.785962 & 0.855498 \\
54 & 0.052603 & 0.707160 & 1.0 & 0.828463 & 0.800728 & 0.865466 \\
55 & 0.054454 & 0.722501 & 1.0 & 0.838897 & 0.815177 & 0.875221 \\
56 & 0.059021 & 0.739263 & 1.0 & 0.850087 & 0.830278 & 0.885417 \\
57 & 0.059345 & 0.756508 & 1.0 & 0.861378 & 0.845117 & 0.895435 \\
58 & 0.061814 & 0.774042 & 1.0 & 0.872631 & 0.859526 & 0.905162 \\
59 & 0.062266 & 0.792979 & 1.0 & 0.884538 & 0.874372 & 0.915185 \\
60 & 0.063552 & 0.813056 & 1.0 & 0.896890 & 0.889357 & 0.925302 \\
61 & 0.064624 & 0.834007 & 1.0 & 0.909492 & 0.904225 & 0.935340 \\
62 & 0.066418 & 0.856123 & 1.0 & 0.922485 & 0.919129 & 0.945402 \\
63 & 0.071635 & 0.879033 & 1.0 & 0.935623 & 0.933779 & 0.955293 \\
64 & 0.080885 & 0.903258 & 1.0 & 0.949170 & 0.948461 & 0.965205 \\
65 & 0.081862 & 0.928700 & 1.0 & 0.963032 & 0.963056 & 0.975058 \\
66 & 0.083409 & 0.956018 & 1.0 & 0.977515 & 0.977862 & 0.985054 \\
\textcolor{red}{67} & \textcolor{red}{0.092559} & \textcolor{red}{0.984360} & \textcolor{red}{1.0} & \textcolor{red}{0.992119} & \textcolor{red}{0.992354} & \textcolor{red}{0.994838} \\
68 & 0.247001 & 1.0 & 0.984945 & 0.992416 & 1.0 & 0.995109 \\
69 & 0.247376 & 1.0 & 0.954177 & 0.976551 & 1.0 & 0.985113 \\
70 & 0.247714 & 1.0 & 0.923385 & 0.960167 & 1.0 & 0.975110 \\
71 & 0.248124 & 1.0 & 0.892670 & 0.943292 & 1.0 & 0.965131 \\
72 & 0.248636 & 1.0 & 0.861909 & 0.925833 & 1.0 & 0.955137 \\
73 & 0.249020 & 1.0 & 0.831072 & 0.907743 & 1.0 & 0.945119 \\
74 & 0.249385 & 1.0 & 0.800288 & 0.889067 & 1.0 & 0.935118 \\
75 & 0.249814 & 1.0 & 0.769550 & 0.869769 & 1.0 & 0.925132 \\
76 & 0.250269 & 1.0 & 0.738720 & 0.849729 & 1.0 & 0.915117 \\
77 & 0.250836 & 1.0 & 0.708035 & 0.829064 & 1.0 & 0.905148 \\
78 & 0.251605 & 1.0 & 0.677221 & 0.807551 & 1.0 & 0.895137 \\
79 & 0.253747 & 1.0 & 0.646392 & 0.785222 & 1.0 & 0.885121 \\
80 & 0.255437 & 1.0 & 0.615608 & 0.762076 & 1.0 & 0.875120 \\
81 & 0.256051 & 1.0 & 0.584832 & 0.738036 & 1.0 & 0.865122 \\
82 & 0.258708 & 1.0 & 0.554063 & 0.713051 & 1.0 & 0.855126 \\
83 & 0.259548 & 1.0 & 0.523325 & 0.687082 & 1.0 & 0.845140 \\
84 & 0.261897 & 1.0 & 0.492457 & 0.659928 & 1.0 & 0.835112 \\
85 & 0.263021 & 1.0 & 0.461757 & 0.631784 & 1.0 & 0.825138 \\
86 & 0.263048 & 1.0 & 0.430829 & 0.602209 & 1.0 & 0.815090 \\
87 & 0.263070 & 1.0 & 0.400455 & 0.571892 & 1.0 & 0.805222 \\
88 & 0.263121 & 1.0 & 0.369103 & 0.539189 & 1.0 & 0.795037 \\
89 & 0.263373 & 1.0 & 0.338592 & 0.505892 & 1.0 & 0.785124 \\
90 & 0.263605 & 1.0 & 0.307823 & 0.470741 & 1.0 & 0.775128 \\
91 & 0.263857 & 1.0 & 0.276766 & 0.433543 & 1.0 & 0.765039 \\
92 & 0.264606 & 1.0 & 0.246240 & 0.395173 & 1.0 & 0.755122 \\
93 & 0.265771 & 1.0 & 0.215487 & 0.354568 & 1.0 & 0.745131 \\
94 & 0.266736 & 1.0 & 0.184703 & 0.311813 & 1.0 & 0.735130 \\
95 & 0.268147 & 1.0 & 0.153896 & 0.266742 & 1.0 & 0.725121 \\
96 & 0.268774 & 1.0 & 0.123112 & 0.219234 & 1.0 & 0.715121 \\
97 & 0.269189 & 1.0 & 0.092306 & 0.169011 & 1.0 & 0.705112 \\
98 & 0.269561 & 1.0 & 0.061568 & 0.115994 & 1.0 & 0.695126 \\
99 & 0.269839 & 1.0 & 0.030708 & 0.059586 & 1.0 & 0.685101 \\
100 & 1.137619 & 1.0 & 0.000030 & 0.000061 & 1.0 & 0.675071 \\
        \bottomrule
    \end{tabular}
    \end{adjustbox}
\label{tab:performance_metrics_51_1003}
\end{table}
\addtocounter{table}{-1}
\begin{table}[h]
    \centering
            \caption{Performance metrics  at different
percentiles of reconstruction errors, demonstrating the trade-offs, and supporting the
selection of the 69th percentile as the optimal threshold for anomaly detection with  65s resampling data.}
    \begin{adjustbox}{width=\textwidth}
    \begin{tabular}{ccccccc}
        \toprule
        Percentile & Optimal Threshold & precision & recall & $\text{F}_1$-score & specificity & accuracy \\
        \midrule
            1.0  & 0.006671  & 0.307389 & 1.0  & 0.470234 & 0.014377 & 0.314317 \\
            2.0  & 0.010666  & 0.310526 & 1.0  & 0.473895 & 0.028751 & 0.324317 \\
            3.0  & 0.012341  & 0.313727 & 1.0  & 0.477614 & 0.043125 & 0.334317 \\
            4.0  & 0.012690  & 0.316995 & 1.0  & 0.481391 & 0.057499 & 0.344317 \\
            5.0  & 0.014052  & 0.320332 & 1.0  & 0.485229 & 0.071873 & 0.354316 \\
            6.0  & 0.014491  & 0.323739 & 1.0  & 0.489129 & 0.086247 & 0.364315 \\
            7.0  & 0.015116  & 0.327220 & 1.0  & 0.493091 & 0.100621 & 0.374315 \\
            8.0  & 0.015249  & 0.330777 & 1.0  & 0.497119 & 0.114995 & 0.384315 \\
            9.0  & 0.016012  & 0.334412 & 1.0  & 0.501212 & 0.129369 & 0.394315 \\
            10.0 & 0.016358  & 0.338128 & 1.0  & 0.505375 & 0.143746 & 0.404317 \\
            11.0 & 0.016491  & 0.341927 & 1.0  & 0.509606 & 0.158120 & 0.414316 \\
            12.0 & 0.017035  & 0.345813 & 1.0  & 0.513909 & 0.172494 & 0.424316 \\
            13.0 & 0.017731  & 0.349788 & 1.0  & 0.518285 & 0.186868 & 0.434316 \\
            14.0 & 0.018500  & 0.353855 & 1.0  & 0.522737 & 0.201242 & 0.444316 \\
            15.0 & 0.019028  & 0.358018 & 1.0  & 0.527265 & 0.215615 & 0.454316 \\
            16.0 & 0.020337  & 0.362280 & 1.0  & 0.531873 & 0.229989 & 0.464315 \\
            17.0 & 0.020939  & 0.366644 & 1.0  & 0.536561 & 0.244363 & 0.474315 \\
            18.0 & 0.021852  & 0.371115 & 1.0  & 0.541334 & 0.258737 & 0.484315 \\
            19.0 & 0.022952  & 0.375698 & 1.0  & 0.546193 & 0.273114 & 0.494317 \\
            20.0 & 0.023814  & 0.380394 & 1.0  & 0.551139 & 0.287488 & 0.504316 \\
            21.0 & 0.024190  & 0.385209 & 1.0  & 0.556175 & 0.301862 & 0.514316 \\
            22.0 & 0.024612  & 0.390148 & 1.0  & 0.561304 & 0.316236 & 0.524316 \\
            23.0 & 0.024765  & 0.395214 & 1.0  & 0.566529 & 0.330610 & 0.534315 \\
            24.0 & 0.025395  & 0.400414 & 1.0  & 0.571851 & 0.344984 & 0.544316 \\
            25.0 & 0.026067  & 0.405753 & 1.0  & 0.577275 & 0.359358 & 0.554315 \\
            26.0 & 0.026479  & 0.411236 & 1.0  & 0.582803 & 0.373732 & 0.564315 \\
            27.0 & 0.027067  & 0.416869 & 1.0  & 0.588437 & 0.388106 & 0.574315 \\
            28.0 & 0.027698  & 0.422660 & 1.0  & 0.594183 & 0.402483 & 0.584316 \\
            29.0 & 0.028432  & 0.428613 & 1.0  & 0.600041 & 0.416857 & 0.594316 \\
            30.0 & 0.028545  & 0.434736 & 1.0  & 0.606015 & 0.431231 & 0.604316 \\
            31.0 & 0.028979  & 0.441036 & 1.0  & 0.612110 & 0.445605 & 0.614315 \\
            32.0 & 0.030082  & 0.447522 & 1.0  & 0.618328 & 0.459979 & 0.624315 \\
            33.0 & 0.030927  & 0.454201 & 1.0  & 0.624675 & 0.474353 & 0.634315 \\
            34.0 & 0.032055  & 0.461083 & 1.0  & 0.631152 & 0.488727 & 0.644315 \\
            35.0 & 0.033005  & 0.468176 & 1.0  & 0.637766 & 0.503100 & 0.654315 \\
            36.0 & 0.033521  & 0.475491 & 1.0  & 0.644519 & 0.517474 & 0.664315 \\
            37.0 & 0.033939  & 0.483040 & 1.0  & 0.651419 & 0.531852 & 0.674316 \\
            38.0 & 0.034578  & 0.490831 & 1.0  & 0.658466 & 0.546226 & 0.684316 \\
            39.0 & 0.036113  & 0.498877 & 1.0  & 0.665668 & 0.560599 & 0.694316 \\
            40.0 & 0.036358  & 0.507192 & 1.0  & 0.673029 & 0.574973 & 0.704316 \\
            41.0 & 0.036784  & 0.515788 & 1.0  & 0.680554 & 0.589347 & 0.714315 \\
            42.0 & 0.036888  & 0.524681 & 1.0  & 0.688250 & 0.603721 & 0.724315 \\
            43.0 & 0.037256  & 0.533885 & 1.0  & 0.696122 & 0.618095 & 0.734315 \\
            44.0 & 0.037454  & 0.543419 & 1.0  & 0.704175 & 0.632469 & 0.744315 \\
            45.0 & 0.037617  & 0.553299 & 1.0  & 0.712418 & 0.646843 & 0.754314 \\
            46.0 & 0.038178  & 0.563547 & 1.0  & 0.720857 & 0.661220 & 0.764316 \\
            47.0 & 0.039061  & 0.574180 & 1.0  & 0.729497 & 0.675594 & 0.774316 \\
            48.0 & 0.039714  & 0.585221 & 1.0  & 0.738347 & 0.689968 & 0.784316 \\
            49.0 & 0.040355  & 0.596696 & 1.0  & 0.747413 & 0.704342 & 0.794315 \\
            50.0 & 0.041110  & 0.608630 & 1.0  & 0.756706 & 0.718716 & 0.804315 \\
        \bottomrule
    \end{tabular}
    \end{adjustbox}
    \label{tab:performance_metrics_1_502}
\end{table}
\addtocounter{table}{-1}
\begin{table}
\renewcommand\thetable{\ref{tab:performance_metrics_1_502}}
    \centering
    \caption{Cont.}
    \begin{adjustbox}{width=\textwidth}
    \begin{tabular}{ccccccc}
        \toprule
        Percentile & Optimal Threshold & precision & recall & $\text{F}_1$-score & specificity & accuracy \\
        \midrule
        51 & 0.041448 & 0.621665 & 1.0 & 0.766699 & 0.733786 & 0.814799 \\
        52 & 0.041923 & 0.634732 & 1.0 & 0.776558 & 0.748272 & 0.824877 \\
        53 & 0.042794 & 0.648075 & 1.0 & 0.786463 & 0.762461 & 0.834747 \\
        54 & 0.043785 & 0.662063 & 1.0 & 0.796677 & 0.776722 & 0.844669 \\
        55 & 0.045243 & 0.676877 & 1.0 & 0.807307 & 0.791182 & 0.854728 \\
        56 & 0.046965 & 0.692268 & 1.0 & 0.818154 & 0.805549 & 0.864724 \\
        57 & 0.047245 & 0.708431 & 1.0 & 0.829335 & 0.819966 & 0.874753 \\
        58 & 0.049649 & 0.725328 & 1.0 & 0.840800 & 0.834350 & 0.884760 \\
        59 & 0.053066 & 0.743171 & 1.0 & 0.852666 & 0.848830 & 0.894833 \\
        60 & 0.058090 & 0.761818 & 1.0 & 0.864809 & 0.863237 & 0.904856 \\
        61 & 0.065950 & 0.780976 & 1.0 & 0.877020 & 0.877323 & 0.914655 \\
        62 & 0.072420 & 0.801347 & 1.0 & 0.889720 & 0.891561 & 0.924561 \\
        63 & 0.075440 & 0.822972 & 1.0 & 0.902891 & 0.905905 & 0.934540 \\
        64 & 0.076202 & 0.845608 & 1.0 & 0.916346 & 0.920133 & 0.944438 \\
        65 & 0.077088 & 0.869489 & 1.0 & 0.930189 & 0.934341 & 0.954322 \\
        66 & 0.082129 & 0.895779 & 1.0 & 0.945025 & 0.949106 & 0.964594 \\
        67 & 0.084271 & 0.922321 & 1.0 & 0.959591 & 0.963159 & 0.974370 \\
        68 & 0.091506 & 0.950524 & 1.0 & 0.974634 & 0.977231 & 0.984160 \\
        \textcolor{red}{69} & \textcolor{red}{0.098479} & \textcolor{red}{0.981761} & \textcolor{red}{1.0} & \textcolor{red}{0.990797} & \textcolor{red}{0.991873} & \textcolor{red}{0.994346} \\
        70 & 0.206688 & 1.0 & 0.985892 & 0.992896 & 1.0 & 0.995707 \\
        71 & 0.207152 & 1.0 & 0.952980 & 0.975924 & 1.0 & 0.985691 \\
        72 & 0.207554 & 1.0 & 0.920135 & 0.958406 & 1.0 & 0.975696 \\
        73 & 0.208043 & 1.0 & 0.887214 & 0.940237 & 1.0 & 0.965678 \\
        74 & 0.208644 & 1.0 & 0.854355 & 0.921458 & 1.0 & 0.955678 \\
        75 & 0.209169 & 1.0 & 0.821548 & 0.902033 & 1.0 & 0.945694 \\
        76 & 0.209650 & 1.0 & 0.788658 & 0.881843 & 1.0 & 0.935685 \\
        77 & 0.210060 & 1.0 & 0.755806 & 0.860922 & 1.0 & 0.925688 \\
        78 & 0.210598 & 1.0 & 0.722908 & 0.839172 & 1.0 & 0.915677 \\
        79 & 0.211333 & 1.0 & 0.690109 & 0.816644 & 1.0 & 0.905695 \\
        80 & 0.213318 & 1.0 & 0.657211 & 0.793153 & 1.0 & 0.895684 \\
        81 & 0.213755 & 1.0 & 0.624253 & 0.768665 & 1.0 & 0.885655 \\
        82 & 0.213862 & 1.0 & 0.591522 & 0.743341 & 1.0 & 0.875694 \\
        83 & 0.214154 & 1.0 & 0.558601 & 0.716798 & 1.0 & 0.865676 \\
        84 & 0.214395 & 1.0 & 0.525628 & 0.689064 & 1.0 & 0.855642 \\
        85 & 0.215216 & 1.0 & 0.492844 & 0.660275 & 1.0 & 0.845665 \\
        86 & 0.215768 & 1.0 & 0.460151 & 0.630278 & 1.0 & 0.835716 \\
        87 & 0.216072 & 1.0 & 0.427071 & 0.598528 & 1.0 & 0.825649 \\
        88 & 0.216426 & 1.0 & 0.394302 & 0.565591 & 1.0 & 0.815677 \\
        89 & 0.217883 & 1.0 & 0.361465 & 0.530995 & 1.0 & 0.805684 \\
        90 & 0.220101 & 1.0 & 0.328583 & 0.494637 & 1.0 & 0.795678 \\
        91 & 0.221149 & 1.0 & 0.295753 & 0.456496 & 1.0 & 0.785687 \\
        92 & 0.222492 & 1.0 & 0.262894 & 0.416335 & 1.0 & 0.775688 \\
        93 & 0.223228 & 1.0 & 0.230041 & 0.374038 & 1.0 & 0.765690 \\
        94 & 0.223757 & 1.0 & 0.197159 & 0.329378 & 1.0 & 0.755683 \\
        95 & 0.224262 & 1.0 & 0.164329 & 0.282273 & 1.0 & 0.745693 \\
        96 & 0.225074 & 1.0 & 0.131416 & 0.232304 & 1.0 & 0.735677 \\
        97 & 0.226338 & 1.0 & 0.098602 & 0.179505 & 1.0 & 0.725691 \\
        98 & 0.227912 & 1.0 & 0.065682 & 0.123267 & 1.0 & 0.715673 \\
        99 & 0.231056 & 1.0 & 0.032867 & 0.063643 & 1.0 & 0.705687 \\
        100 & 1.103892 & 0.0 & 0.0 & 0.0 & 1.0 & 0.695685 \\
        \bottomrule
    \end{tabular}
    \end{adjustbox}

\label{tab:performance_metrics_51_1002}
\end{table}

\section{}
\label{apendixB}

We categorized all false anomalies identified by models trained and validated on the three datasets. The categorization process involved identifying false anomalies that resemble each of the three predefined anomaly criteria.

Figure \ref{Piechart} shows the distribution of false anomalies for the datasets without resampling (a), with 90s resampling (b), and with 65s resampling (c), respectively.  Red indicates false anomalies that do not resemble any predefined anomaly criteria (classified as unknown). 
Yellow are the false anomalies that resemble anomaly type \textit{fish recorded at only one station} (Criterion 1), green are false anomalies that resemble anomaly of type \textit{normal/expected movements, but then remained stationary for more than 120 days}   (Criterion 2), and blue are false anomalies that resemble anomaly type \textit{a fish passing more than one consecutively positioned receiver} (Criterion 3). The results show that some false anomalies closely resembled true anomalies falling under type \textit{a fish passing more than one consecutively positioned receiver}, while others did not resemble any of the predefined criteria. 

\begin{figure}
    \centering
    \begin{subfigure}[t]{0.3\textwidth}
        \centering
        \includegraphics[width=1.1\textwidth]{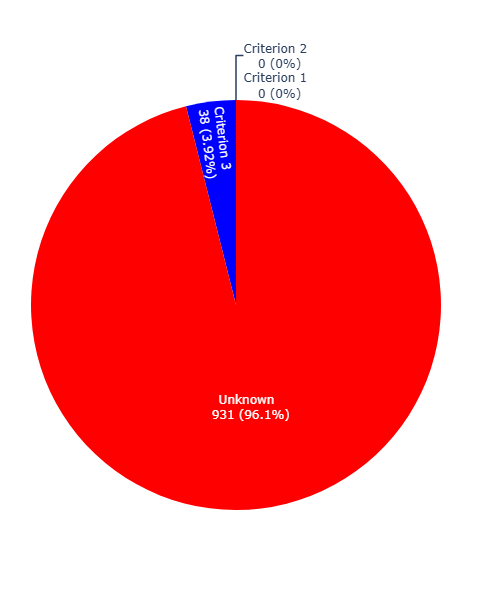}
        \caption{}
    \end{subfigure}
    \begin{subfigure}[t]{0.3\textwidth}
        \centering
        \includegraphics[width=1.1\textwidth]{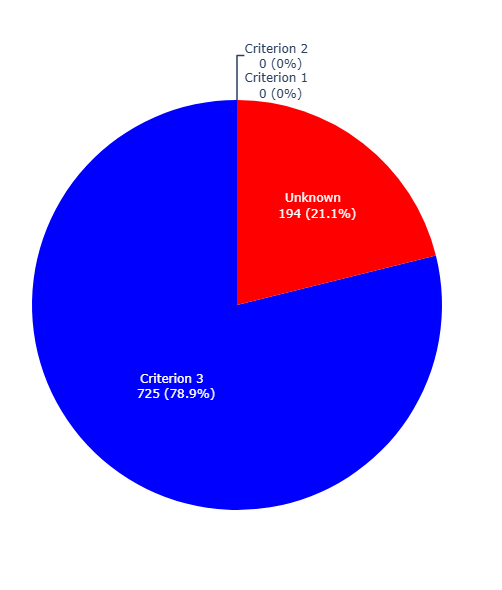}
        \caption{}
    \end{subfigure}   
    \begin{subfigure}[t]{0.3\textwidth}
        \centering
        \includegraphics[width=1.4\textwidth]{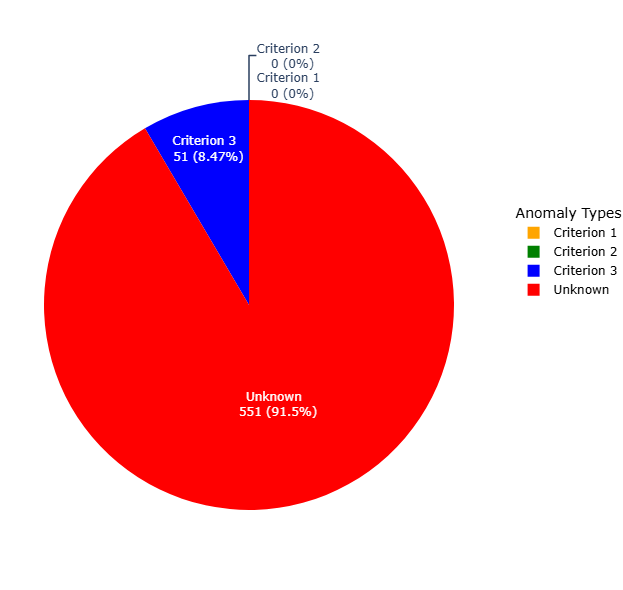}
        \caption{}
    \end{subfigure}
    \caption{(a) False anomaly distribution for without sampling, (b) false anomaly distribution for 90s sampling interval, and (c) false anomaly distribution for 65s sampling interval.}
    \label{Piechart}
\end{figure}

\section{}
\label{Apendic}
Performance comparison of classifiers without resampling and with resampling at 90s and 65s intervals. Metrics evaluated include accuracy, recall, specificity, precision, $\text{F}_1$-score, AUC, and the number of parameters and training time, as shown in Table \ref{tab:classifiers}. A visual representation of these metrics is provided in Figure \ref{fig:Model_Evaluation_Metrics}.
\begin{landscape}
\begin{table}[h]
\centering
\caption{Performance comparison of classifiers without resampling and with resampling at 90s and 65s intervals. Metrics evaluated include accuracy, recall, specificity, precision, $\text{F}_1$-score, AUC, and the number of parameters and training time.}
\scalebox{0.65}{
\begin{tabular}{|c|c|c|c|c|c|c|c|c|c|c|c|c|}
\hline
\textbf{Classifiers} & \multicolumn{3}{c|}{\textbf{IF}} & \multicolumn{3}{c|}{\textbf{DBSCAN}} & \multicolumn{3}{c|}{\textbf{LOF}} & \multicolumn{3}{c|}{\textbf{NN-AE}} \\ \hline
\textbf{Resampling} & \textbf{NO} & \multicolumn{2}{c|}{\textbf{YES}} & \textbf{NO} & \multicolumn{2}{c|}{\textbf{YES}} & \textbf{NO} & \multicolumn{2}{c|}{\textbf{YES}} & \textbf{NO} & \multicolumn{2}{c|}{\textbf{YES}} \\ 
 &  & \textbf{90s} & \textbf{65s} &  & \textbf{90s} & \textbf{65s} &  & \textbf{90s} & \textbf{65s} &  & \textbf{90s} & \textbf{65s} \\ \hline
\textbf{Accuracy} & 67.60$\pm$0.0011 & 67.61$\pm$ 0.0012 &67.62$\pm$0.0012 & 66.00$\pm$0.0011 & 66.00$\pm$0.0012 & 66.00$\pm$0.0012 & 66.83$\pm$0.0013 & 66.83$\pm$0.0012 & 66.83$\pm$0.0013 &   99.76$\pm$0.0013 &  99.77$\pm$0.0002 & 99.85$\pm$0.0002 \\ 
\textbf{Recall} & 0.10$\pm$0.0001 &  0.15$\pm$ 0.0002 & 0.16$\pm$ 0.0002 & 0.24$\pm$0.0002 & 0.24$\pm$0.0002 & 0.24$\pm$0.0002 & 0.40$\pm$0.0003 & 0.40$\pm$0.0003 & 0.40$\pm$0.0003 & 100$\pm$0.0000 & 100$\pm$0.0000 & 100$\pm$0.0000 \\ 
\textbf{Specificity} & 100$\pm$0.0000 & 100$\pm$0.0000 & 100$\pm$0.0000 & 97.57$\pm$0.0002  & 97.57$\pm$0.0005 & 97.57$\pm$0.0004  &  98.71$\pm$0.0004 & 98.71$\pm$0.0003 & 98.71$\pm$0.0003 & 99.64$\pm$0.0002  & 99.66$\pm$0.0003 &  99.78$\pm$0.0003\\ 
\textbf{Precision} & 100$\pm$0.0000 & 100$\pm$0.0000 & 100$\pm$0.0000 & 4.69$\pm$0.0036 & 4.69$\pm$0.0040 & 4.69$\pm$0.0040 & 13.16$\pm$0.0092 & 13.16$\pm$0.0091 & 13.16$\pm$0.0084  & 99.27$\pm$0.0092 & 99.31$\pm$0.0005 & 99.55$\pm$0.0006 \\ 
\textbf{$\text{F}_1$-score} &0.21$\pm$0.0003 & 0.30$\pm$0.0003 & 0.33$\pm$0.0004 & 0.47$\pm$0.0004 & 0.47$\pm$0.0004 &0.47$\pm$0.0005 & 0.78$\pm$0.0006 & 0.78$\pm$0.0005 & 0.78$\pm$0.0006  & 99.63$\pm$0.0005 & 99.65$\pm$0.0002 & 99.77$\pm$0.0003 \\ 
\textbf{AUC} &50.05$\pm$0.0001 & 50.07$\pm$0.0001 & 50.08$\pm$0.0001 & 48.90$\pm$0.0003 & 48.90$\pm$0.0003 &48.90$\pm$0.0002  & 49.56$\pm$0.0003 & 49.56$\pm$0.0002 & 49.56$\pm$0.0002 & 99.82$\pm$0.0001 & 99.83$\pm$0.0001 & 99.89$\pm$0.0001 \\ 
\textbf{No. of parameters} & 5 & 5 & 5& 2 & 2 & 2  &5  & 5 &  5& 3597 & 3597 & 3597 \\ 
\textbf{Train time (minutes)} & 4.17 &  4.55 & 4.74&8.92 &9.13 & 9.44 & 24.61 & 26.73 & 32.48  & 699.65 & 737.97 & 852.87\\ \hline
\end{tabular}
}
\label{tab:classifiers}
\end{table}
\end{landscape}

\begin{figure}
    \centering
    \includegraphics[width=\textwidth]{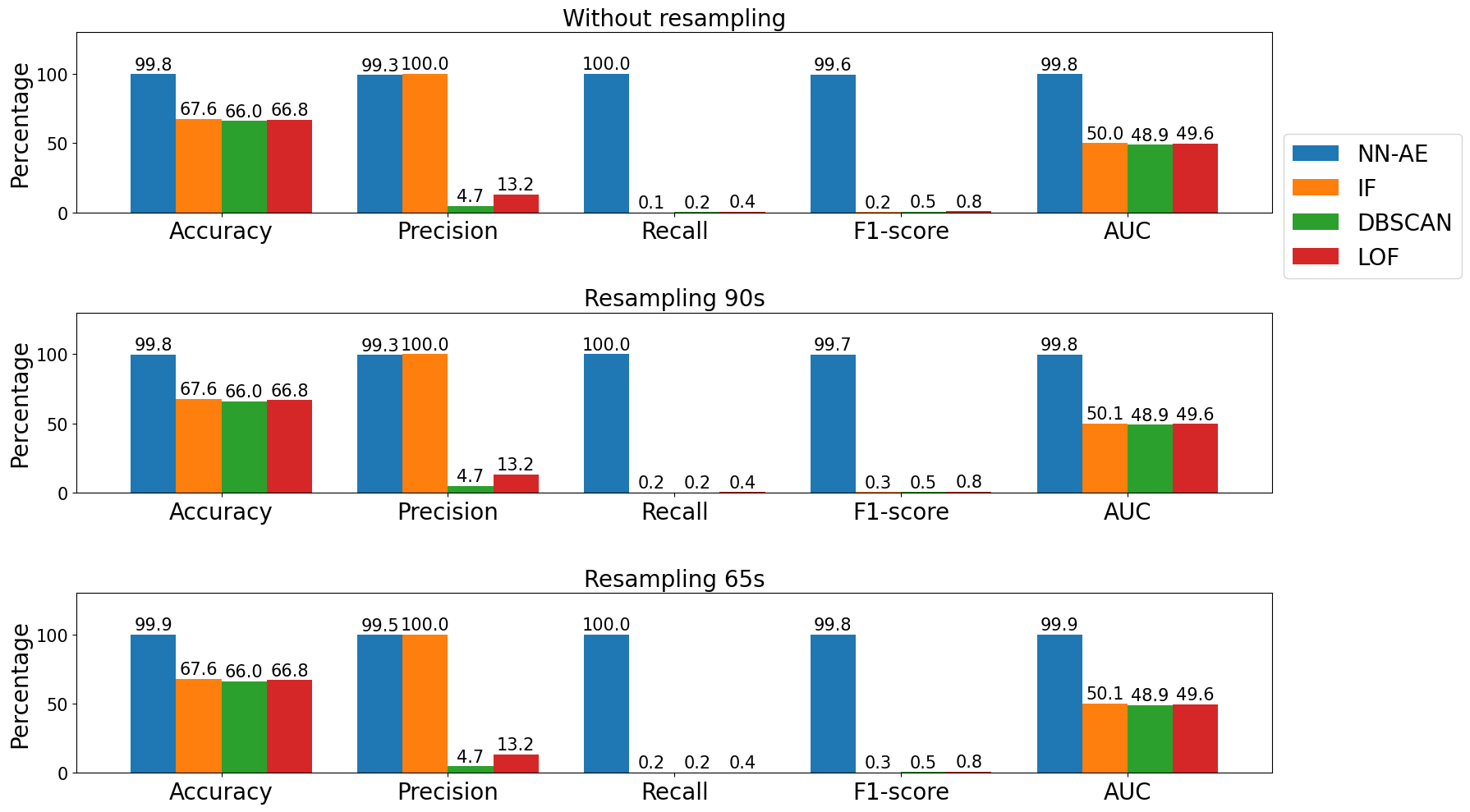}
    \caption{This bar chart shows the key performance metrics (Precision, Recall, Accuracy, $\text{F}_1$-score, and AUC) for different models (NN-AE, IF, DBSCAN, LOF) evaluated on data  without resampling, with 90s resampling, and with 65s resampling. The NN-AE model consistently achieves near-perfect performance, significantly outperforming the traditional models.}
    \label{fig:Model_Evaluation_Metrics}
\end{figure}

\section{}
\label{Apend:MVM}
{Figures \ref{fig:move1}, \ref{fig:move2}, \ref{fig:move3}, and \ref{fig:move4} illustrate the movement patterns of adult dusky kob in the Breede Estuary, highlighting anomalies and TA as detected by the NN-AE, IF, LOF, and DBSCAN algorithms. Each figure consists of two parts: (a) provides an overview of dusky kob movements from 2016 to 2021, represented by blue lines indicating latitudinal shifts over time. Red dots denote anomalies and magenta crosses indicate TA, where the models correctly identified unusual movements.}

{At the top of each panel (a), a black square zoomed a specific region that focuses on a particular time frame in 2020. This region is examined in more detail in panel (b) of each figure, which zooms in on fish movements during this period. The zoomed-in view provides a granular analysis of daily or monthly fish movements, enabling a closer evaluation of how each model performs in detecting anomalies. By narrowing the focus to this specific interval, the detailed visualization helps assess the models' precision and accuracy in identifying TA, offering insights into their reliability and effectiveness in capturing significant behavioural deviations.}
\begin{figure}
    \centering
    
    \begin{minipage}{\textwidth}
        \centering
        \includegraphics[width=0.93\textwidth]{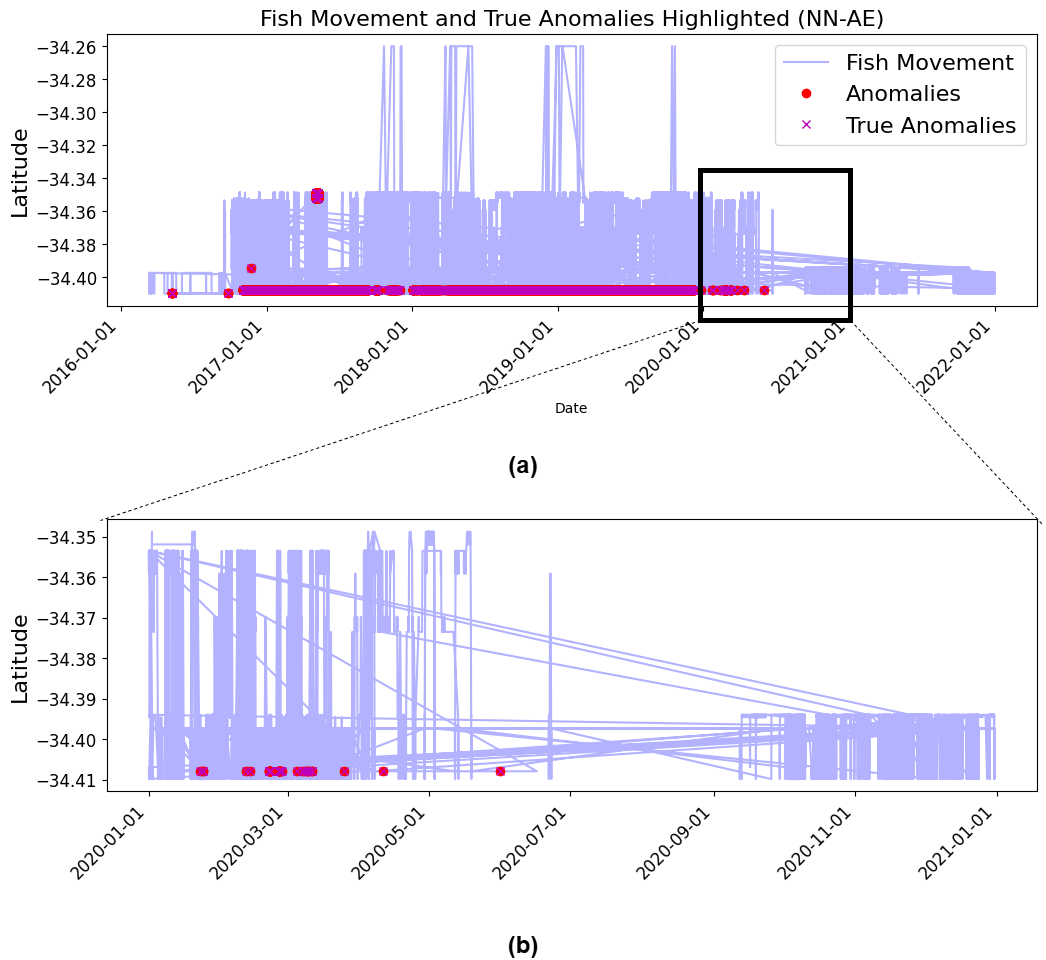}
    \end{minipage}

    \caption{This figure shows the movement patterns of dusky kob in the Breede Estuary from 2016 to 2021, with a focus on highlighting anomalies and TA detected by the NN-AE. (a) An overview from 2016 to 2021, with blue lines indicating latitudinal movements, red dots highlighting anomalies, and magenta crosses marking TA. (b) A zoomed-in view of the 2020 movement patterns, shows a closer examination of anomalies and TA.}

    \label{fig:move1}
\end{figure}

\begin{figure}
    \centering
    
    \begin{minipage}{\textwidth}
        \centering
        \includegraphics[width=0.93\textwidth]{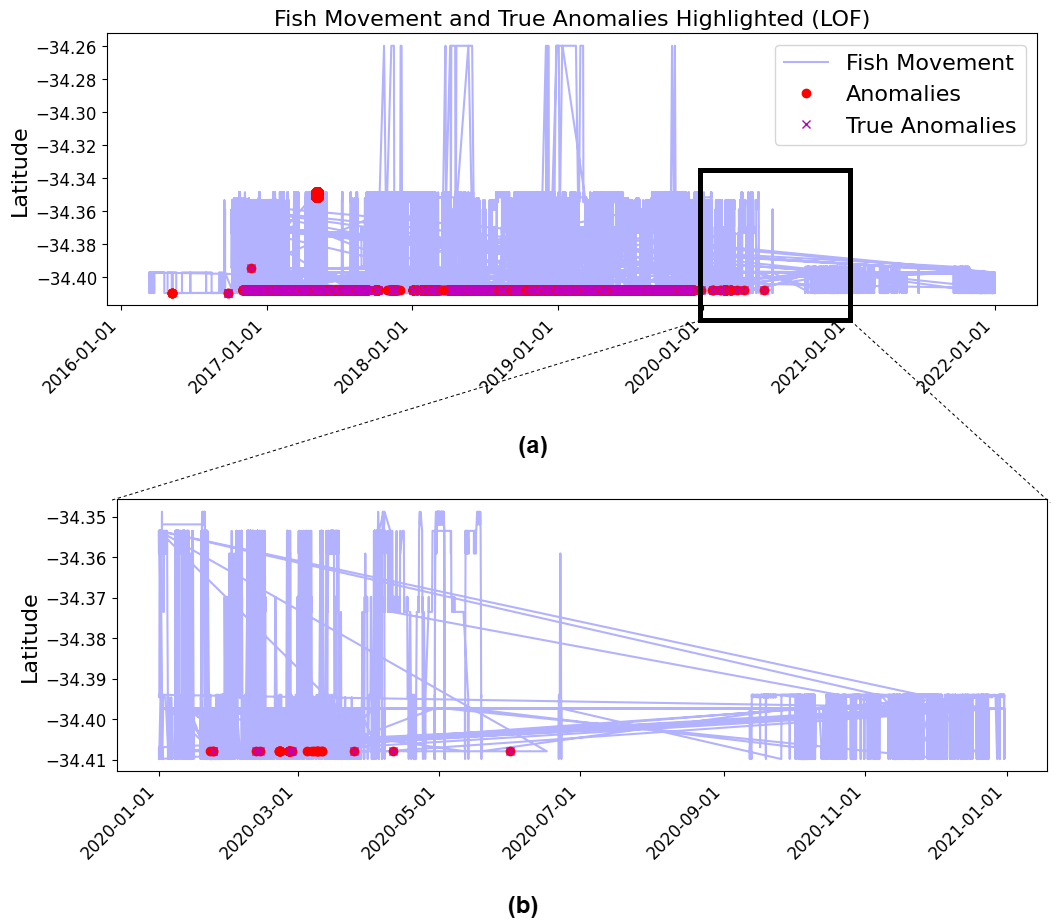}
    \end{minipage}

    \caption{This figure shows the movement patterns of dusky kob in the Breede Estuary from 2016 to 2021, with a focus on highlighting anomalies and TA detected by the LOF. (a) An overview from 2016 to 2021, with blue lines indicating latitudinal movements, red dots highlighting anomalies, and magenta crosses marking TA. (b) A zoomed-in view of the 2020 movement patterns, shows a closer examination of anomalies and TA.}
    \label{fig:move2}
\end{figure}

\begin{figure}
    \centering
    
    \begin{minipage}{\textwidth}
        \centering
        \includegraphics[width=0.93\textwidth]{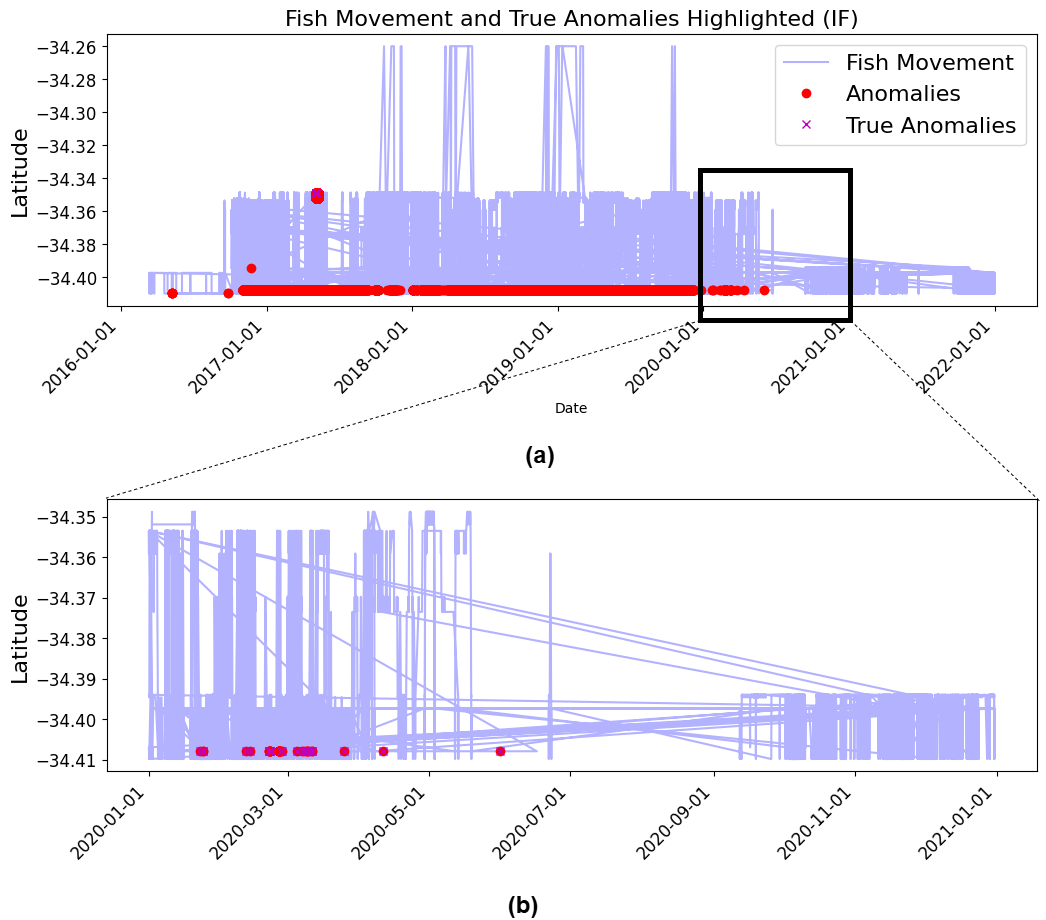}
    \end{minipage}

    \caption{This figure shows the movement patterns of dusky kob in the Breede Estuary from 2016 to 2021, with a focus on highlighting anomalies and TA detected by the IF. (a) An overview from 2016 to 2021, with blue lines indicating latitudinal movements, red dots highlighting anomalies, and magenta crosses marking TA. (b) A zoomed-in view of the 2020 movement patterns, shows a closer examination of anomalies and TA.}
    \label{fig:move3}
\end{figure}

\begin{figure}
    \centering
    
    \begin{minipage}{\textwidth}
        \centering
        \includegraphics[width=0.93\textwidth]{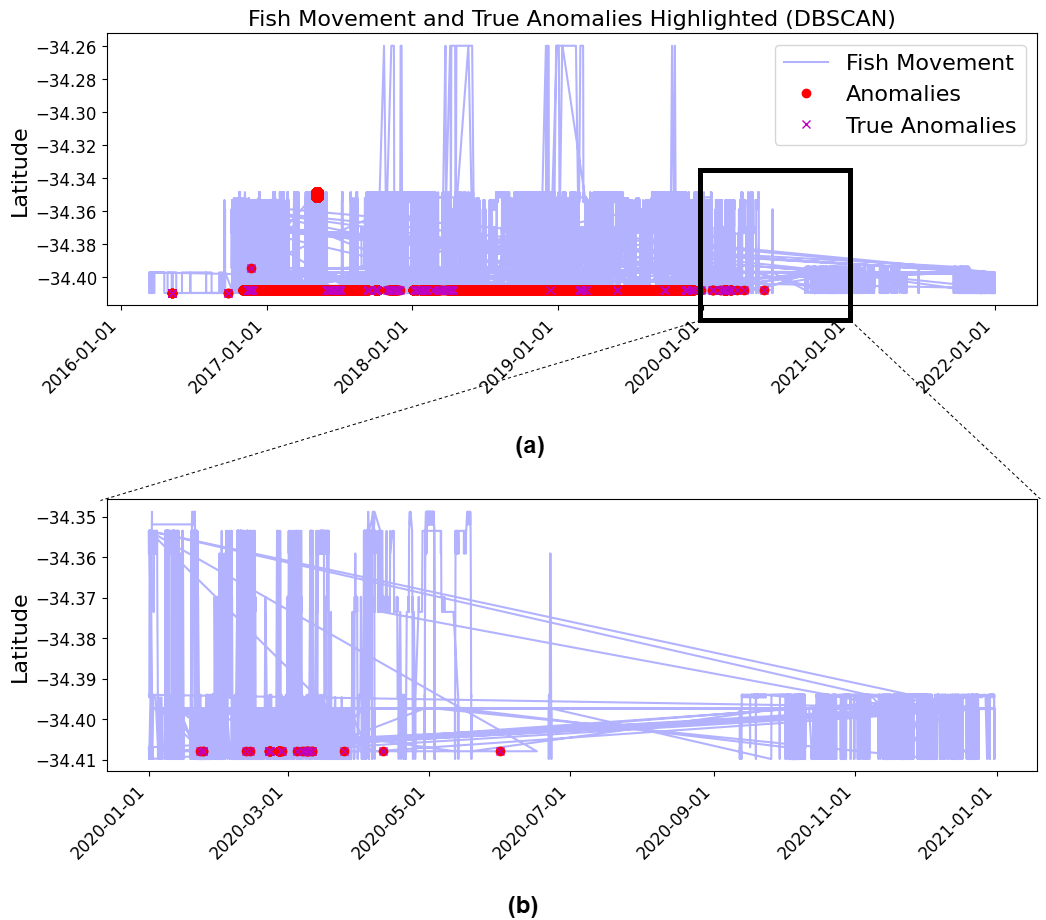}
    \end{minipage}

    \caption{This figure shows the movement patterns of dusky kob in the Breede Estuary from 2016 to 2021, with a focus on highlighting anomalies and TA detected by the DBSCAN. (a) An overview from 2016 to 2021, with blue lines indicating latitudinal movements, red dots highlighting anomalies, and magenta crosses marking TA. (b) A zoomed-in view of the 2020 movement patterns, shows a closer examination of anomalies and TA.}
    \label{fig:move4}
\end{figure}

{Figure \ref{fig:FP} compares the performance of the NN-AE, IF, DBSCAN, and LOF models in FA. Panel (b) of Figure \ref{fig:FP} shows that the IF model does not produce any FA, while NN-AE generates the fewest FA among the remaining models, followed by LOF and DBSCAN. This indicates that IF is the most effective in avoiding the misclassification of normal behaviours as anomalies.}

{However, the absence of FA does not necessarily make IF the best overall model. While NN-AE produces slightly more FA than IF, it achieves 100\% of TA with negligible FA. NN-AE's capacity to learn complex spatial and temporal patterns enables it to identify meaningful anomalies that reflect significant changes in fish behaviour or environmental conditions, without overrepresenting irrelevant data. This capability positions NN-AE as the most suitable model for anomaly detection in telemetry data.}

\begin{figure}
    \centering
    \begin{subfigure}[t]{\textwidth}
        \centering
        \includegraphics[width=0.78\textwidth]{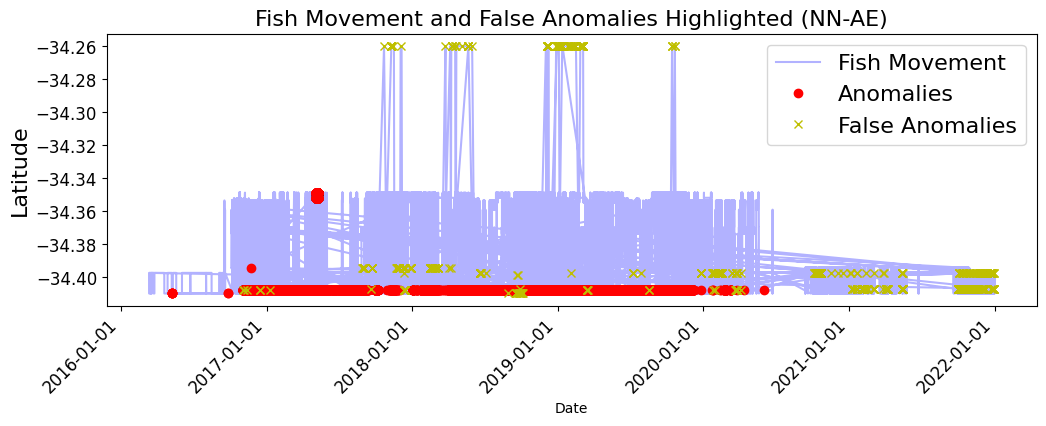}
        \caption{}
    \end{subfigure}\\

    \begin{subfigure}[t]{\textwidth}
        \centering
        \includegraphics[width=0.78\textwidth]{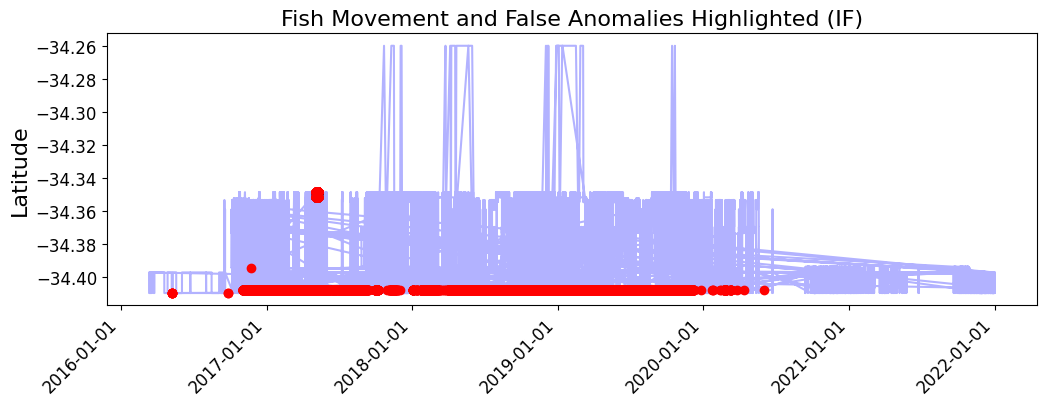}
        \caption{}
    \end{subfigure}\\

    \begin{subfigure}[t]{\textwidth}
        \centering
        \includegraphics[width=0.78\textwidth]{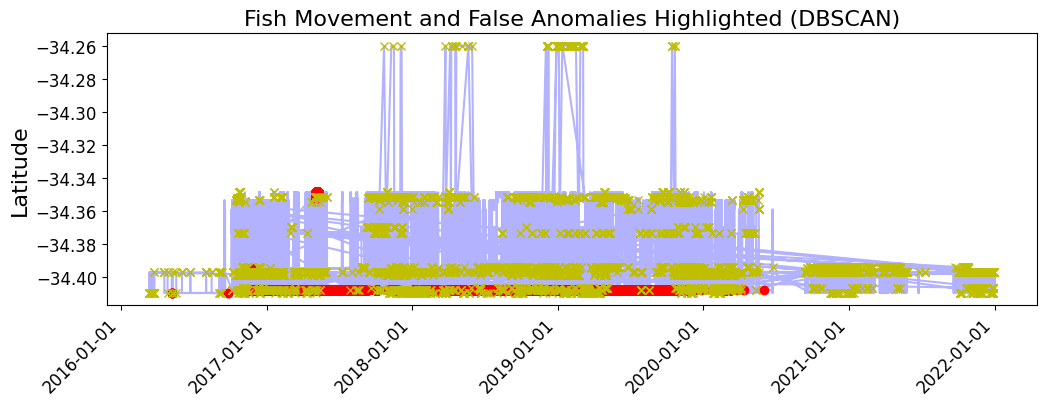}
        \caption{}
    \end{subfigure}\\

    \begin{subfigure}[t]{\textwidth}
        \centering
        \includegraphics[width=0.78\textwidth]{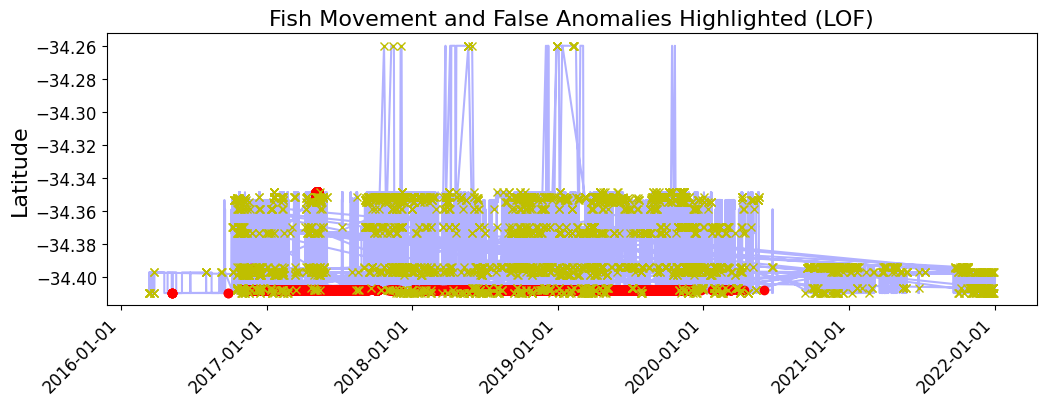}
        \caption{}
    \end{subfigure}

    \caption{This figure shows the movement patterns of dusky kob in the Breede Estuary from 2016 to 2021, highlighting anomalies and FA detected by four machine learning models: NN-AE (a), IF (b), DBSCAN (c), and LOF (d). Blue lines represent fish movements, red dots indicate unusual movement patterns (anomalies) and magenta crosses mark correctly identified FA.}
    \label{fig:FP}
\end{figure}

{Figure \ref{fig:FN} compares fish movement data over time, highlighting anomalies and FN detections. Each plot visualizes the changes in fish latitude from 2016 to 2021, with anomalies marked in red and FN in green. NN-AE model (a) achieves zero FN demonstrating its exceptional accuracy in detecting anomalies. This superior performance is attributed to the model's ability to learn complex temporal and spatial patterns in the fish movement data. The absence of FN indicates that NN-AE effectively identifies all anomalies in the telemetry observation. This capability is particularly valuable for telemetry research, as removing anomalies could provides a more complete understanding of the factors influencing fish movement. In contrast, the IF (b), DBSCAN (c), and LOF (d) models misclassify a significant number of anomalies as normal instances, which undermines their effectiveness in detecting true behavioural anomalies.}

\begin{figure}
    \centering
    \begin{subfigure}[t]{\textwidth}
        \centering
        \includegraphics[width=0.78\textwidth]{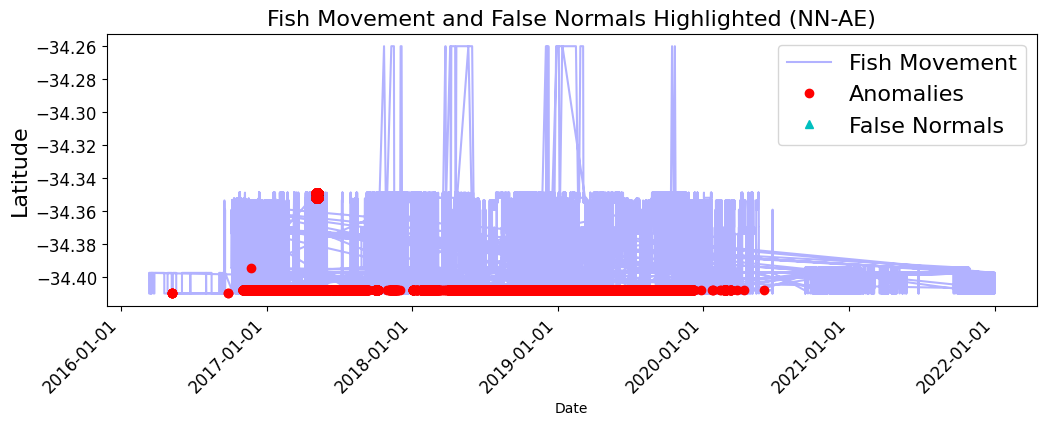}
        \caption{}
    \end{subfigure}\\

    \begin{subfigure}[t]{\textwidth}
        \centering
        \includegraphics[width=0.78\textwidth]{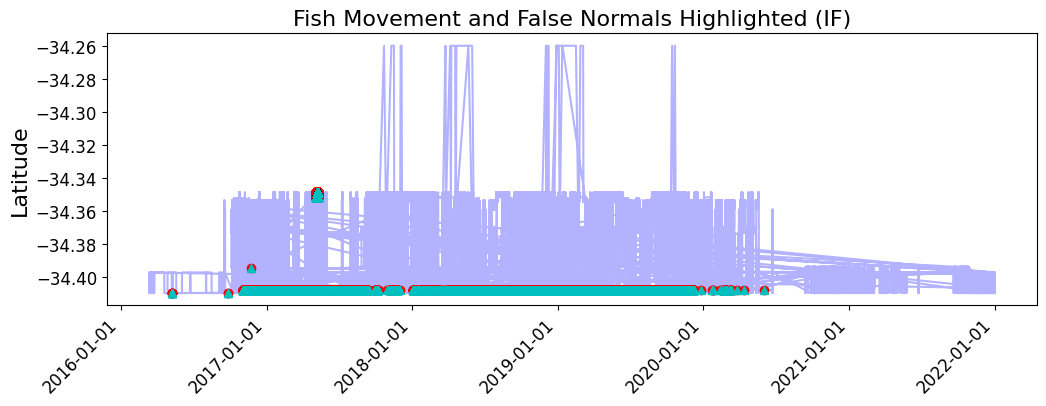}
        \caption{}
    \end{subfigure}\\

    \begin{subfigure}[t]{\textwidth}
        \centering
        \includegraphics[width=0.78\textwidth]{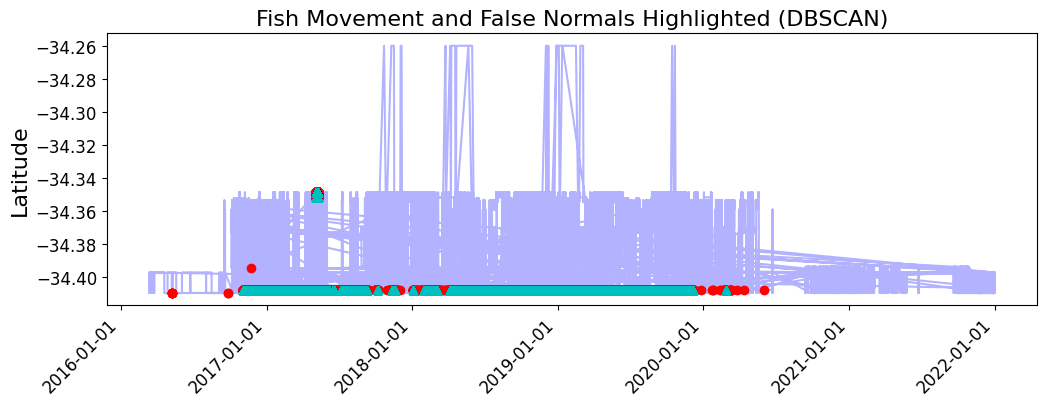}
        \caption{}
    \end{subfigure}\\

    \begin{subfigure}[t]{\textwidth}
        \centering
        \includegraphics[width=0.78\textwidth]{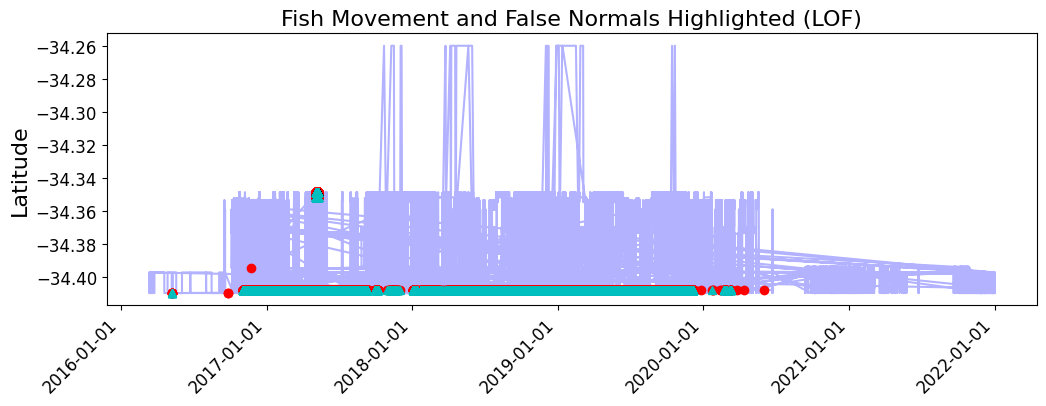}
        \caption{}
    \end{subfigure}

    \caption{This figure shows the movement patterns of dusky kob in the Breede Estuary from 2016 to 2021, highlighting anomalies and FN detected by four machine learning models: NN-AE (a), IF (b), DBSCAN (c), and LOF (d). Blue lines represent fish movements, red dots indicate unusual movement patterns (anomalies) and magenta crosses mark correctly identified FN.}
    \label{fig:FN}
\end{figure}

\begin{figure}
    \centering
    
    \begin{subfigure}[t]{\textwidth}
        \centering
        \includegraphics[width=0.78\textwidth]{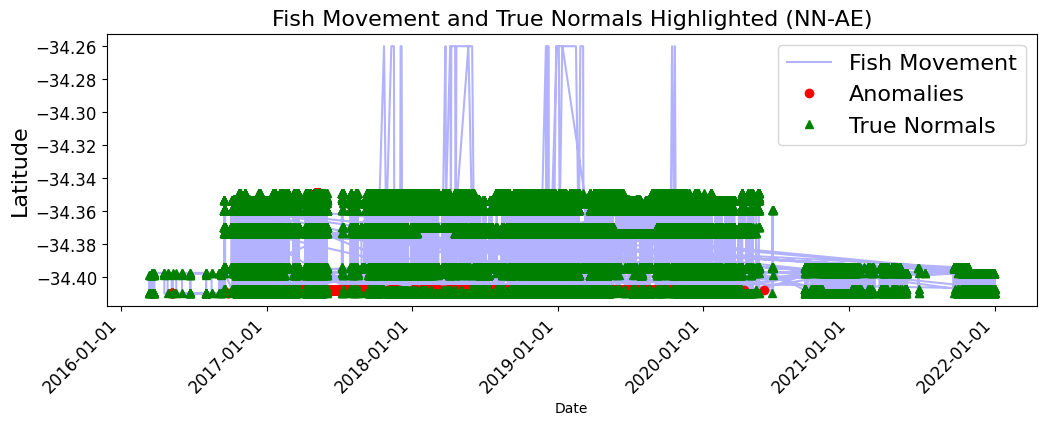}
        \caption{}
    \end{subfigure}\\
    
    \begin{subfigure}[t]{\textwidth}
        \centering
        \includegraphics[width=0.78\textwidth]{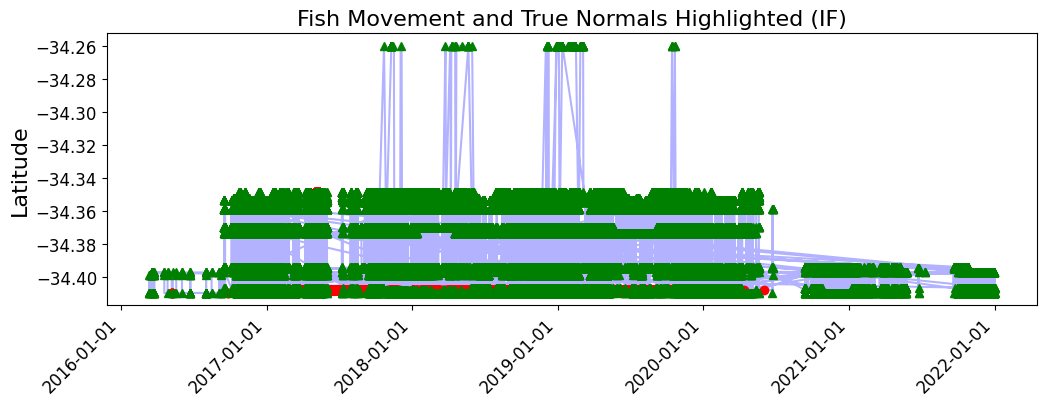}
        \caption{}
    \end{subfigure}\\
    
    \begin{subfigure}[t]{\textwidth}
        \centering
        \includegraphics[width=0.78\textwidth]{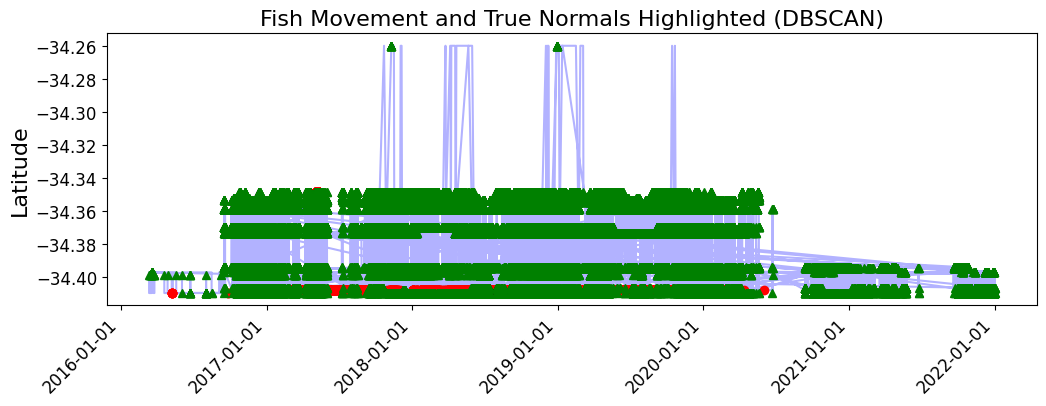}
        \caption{}
    \end{subfigure}\\
    
    \begin{subfigure}[t]{\textwidth}
        \centering
        \includegraphics[width=0.78\textwidth]{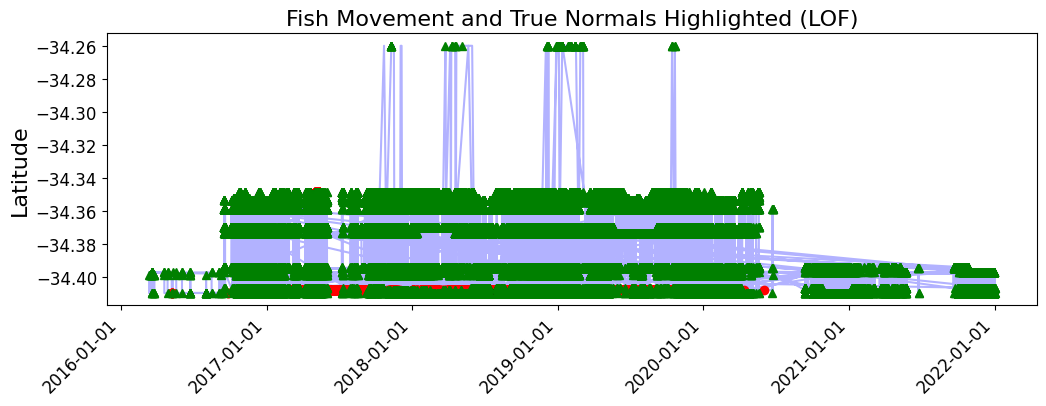}
        \caption{}
    \end{subfigure}
    \caption{This figure shows the movement patterns of dusky kob in the Breede Estuary from 2016 to 2021, highlighting anomalies and TN detected by four machine learning models: NN-AE (a), IF (b), DBSCAN (c), and LOF (d). Blue lines represent fish movements, red dots indicate unusual movement patterns (anomalies) and magenta crosses mark correctly identified TN.}
    \label{fig:TN}
\end{figure}
\end{appendices}
\newpage
\bibliography{sn-article}
\end{document}